\documentclass[10pt,journal,compsoc]{IEEEtran}
\ifCLASSOPTIONcompsoc
  \usepackage[noadjust]{cite}
\else
  \usepackage[noadjust]{cite}
\fi
\ifCLASSINFOpdf
  \usepackage[pdftex]{graphicx}
  \graphicspath{{../images/}{../figs/}}
  \DeclareGraphicsExtensions{.pdf,.jpeg,.png}
\else
\fi
\usepackage{amsmath}
\usepackage{amssymb}
\usepackage{algorithm}
\usepackage{algpseudocode}
\usepackage{url}

\usepackage{color}
\newcommand{\huy}[1]{\textcolor{black}{#1}}

\begin{document}
%
\title{Color Orchestra: Ordering Color Palettes for Interpolation and Prediction}

%
%
%
%

\author{Huy~Q.~Phan,%
        ~Hongbo~Fu,%
        ~and~Antoni~B.~Chan
\IEEEcompsocitemizethanks{
\IEEEcompsocthanksitem H Q. Phan is with the Deparment of Computer Science, University of Bath, UK.
\IEEEcompsocthanksitem A. B. Chan is with the Department of Computer Science, City University of Hong Kong, Hong Kong.
\IEEEcompsocthanksitem H. Fu is with the School of Creative Media, City University of Hong Kong, Hong Kong.}
}

%
%

\markboth{IEEE Transactions on Visualization and Computer Graphics, March~2017}%
{Phan \MakeLowercase{\textit{et al.}}: Color Orchestra: Modeling Palette Density for Photo-style Exploration and Color Suggestion}
%



\IEEEtitleabstractindextext{%
\begin{abstract}~
Color theme or color palette can deeply influence the quality and the feeling of a photograph or a graphical design. Although color palettes may come from different sources such as online crowd-sourcing, photographs and graphical designs, in this paper, we consider color palettes extracted from fine art collections, which we believe to be an abundant source of stylistic and unique color themes. We aim to capture color styles embedded in these collections by means of statistical models and to build practical applications upon these models. As artists often use their personal color themes in their paintings, making these palettes appear frequently in the dataset, we employed \emph{density estimation} to capture the characteristics of palette data. Via density estimation, we carried out various predictions and interpolations on palettes, which led to promising applications such as photo-style exploration, real-time color suggestion, and enriched photo recolorization. It was, however, challenging to apply density estimation to palette data as palettes often come as unordered sets of colors, which make it difficult to use conventional metrics on them. To this end, we developed a divide-and-conquer sorting algorithm to rearrange the colors in the palettes in a coherent order, which allows meaningful interpolation between color palettes. To confirm the performance of our model, we also conducted quantitative experiments on datasets of digitized paintings collected from the Internet and received favorable results.

\end{abstract}

}

\maketitle

\IEEEdisplaynontitleabstractindextext

%
\IEEEpeerreviewmaketitle

 \section{Introduction}\label{smartpalette:sec:intro}
\begin{figure*}
\includegraphics[height=1.9in]{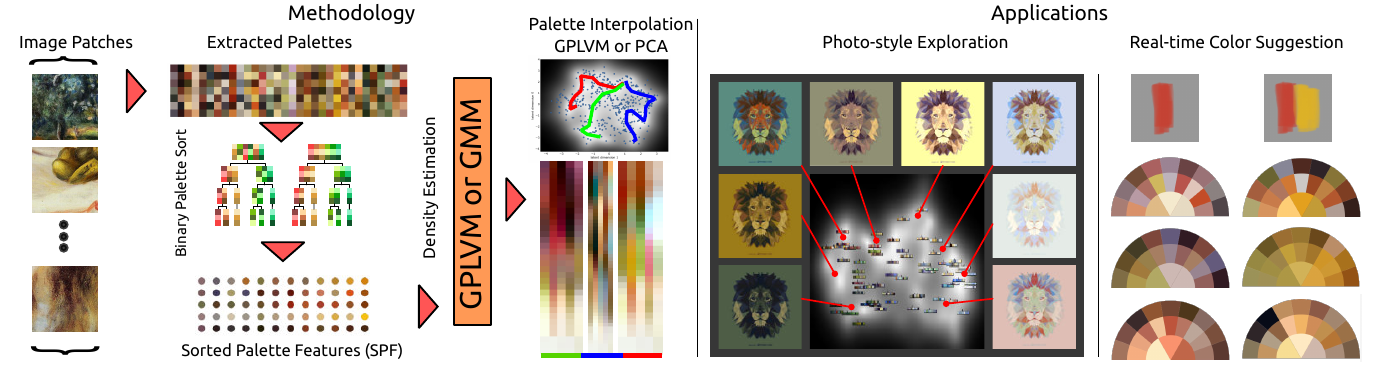}
\caption[SmartPalette: An overview]{An overview of our work. We first extracted color palettes (themes) from image patches, which were uniformly sampled from paintings of one or more artists. Second, our Binary Palette Sort algorithm sorted unordered color palettes. Third, GPLVM was used for palette interpolation and prediction. The last columns show two applications of our method:  photo-style exploration and adaptive palette.}\label{fig:teaser}
\vspace{-10pt}
\end{figure*}
Color is a topic that lies in the intersection of art and science. A large body of literature has been devoted to study color from very different perspectives. Computer scientists have been working on topics such as color transfer, color harmony, grey-scale photo colorization, and color picking interfaces. A relatively new trend in color harmony study is to discover harmonious color combinations from existing color themes, for example, through data-driven approaches. O'Donovan et al.~collect large datasets of color themes along with their ratings and train a regressor to predict ratings for novel color themes~\cite{o2011color}. Lin et al.~later integrate this regressor into a factor graph, which is capable of suggesting interesting colorings for segmented patterns~\cite{lin2013probabilistic}. Huang et al.~employ a similar factor graph for image recolorization~\cite{huang2014learning}. Wang et al.~introduce techniques to enhance color themes with data~\cite{wang2010data,wang2011example}.

Interestingly, few of these works have attempted to model the \textit{density} of palette data, which tells how frequently a certain palette appears in the data. One of the main challenges to modeling palette density is that palette datasets are available as sets of \textit{unordered colors}, making it difficult to directly apply traditional density estimation methods. Representative features introduced in \cite{o2011color} might be used for density estimation but one will lose the ability to interpolate new color palettes. Palette interpolation is potentially useful for applications such as example-based color suggestion and image recolorization (see Section \ref{sec:apps}). When the number of examples is limited, interpolation helps extending the range of available choices possibly infinitely. 

\begin{figure}[ht]
\centering
\includegraphics[width=0.85\linewidth]{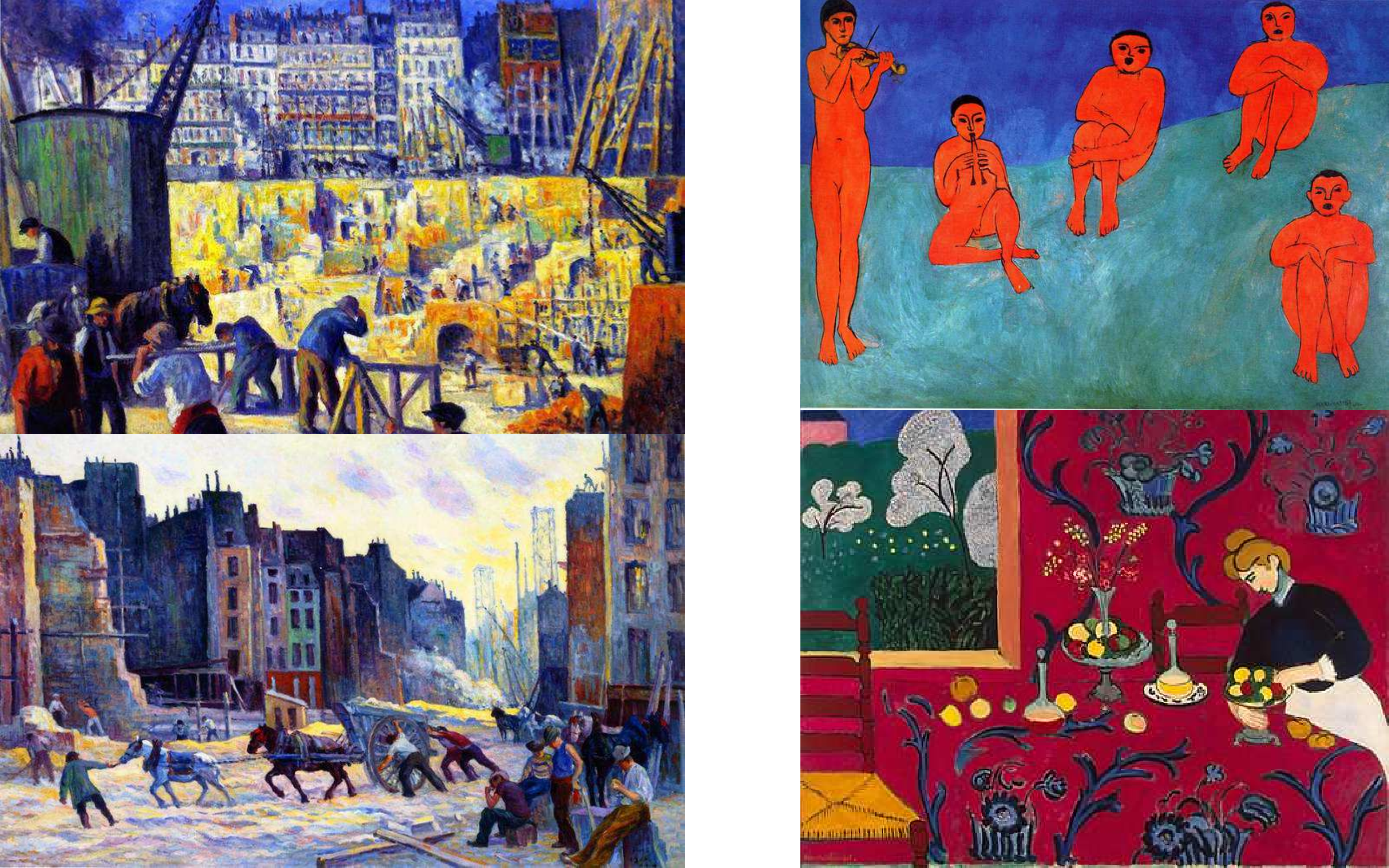}
\caption{(Left) Maximilien Luce's paintings. (Right) Henri Matisse's paintings.}\label{fig:luce_matisse}
\vspace{-15pt}
\end{figure}

A key observation behinds our approach is that color palettes extracted from an art collection often conform to a certain aesthetic style, which can be revealed by their density distribution. This can be observed from paintings of the world-renowned painters such as Paul Cezanne, Vincent van Gogh,  Pierre-Auguste Renoir. It is commonly known that there are certain rules of using colors in fine art~\cite{mollica2013color}. For instance, to color an object, one should use warm colors for illuminated regions and cold colors for shaded regions. Although artists have  their own choices of colors, basic rules are usually respected as they help ensuring that the paintings are perceived properly (e.g., a morning scene should be understood as in a morning time).  On the other hand, distinguished artists often prefer  to use their favorite color combinations over the actual colors of the objects being painted. This is because a repeated use of some sets of colors will eventually make their paintings stand out. For instance, Maximilien Luce used strong blues and yellows to shade objects (Figure~\ref{fig:luce_matisse}); Henri Matisse often used deep reds and greens in his paintings. These observations suggest that high frequency color themes can represent the style of a palette dataset, and with the help of interpolation, we can generate more color palettes with the same style by sampling from a palette density.

To perform palette density estimation and palette interpolation, we impose an order on the colors in a palette dataset so that the Euclidean distance between the palettes correctly reflects the difference between them, and corresponding colors in the palettes are aligned. Our approach achieves this by globally sorting the colors in the dataset so that the total distance between all pairs of color palettes is minimized. The sorting algorithm \---\ called Binary Palette Sort \---\ is one of our main contributions as it greatly improves the performance of many applications introduced in our work. We have thoroughly evaluated its performance with both quantitative and qualitative experiments (see Section~\ref{sec:evals}).
The order of the colors allows us to apply conventional  techniques to estimate the density of color palettes and to interpolate among them. Furthermore, given a palette with missing colors, we can predict compatible colors by sampling from data points that simultaneously have high likelihood and are similar to the input palette. In our implementation, we used Gaussian process latent variable models (GPLVM) to estimate the density of color palettes and to interpolate new palettes. 

Based on palette density estimation and palette interpolation, we developed various applications that would only be possible with these techniques. We introduce a novel application called Photo-style Explorer, which acts like an ``Instagram'' filter but in a continuous design space (Figure~\ref{fig:teaser}). That is, instead of choosing among a few predefined themes, one can freely surf a continuous space of possible photo colorizations to pick a preferred theme. Since the space of color themes is usually of high dimensionality, we utilize dimensionality reduction (GPLVM) to represent it with a familiar 2-D interface. By moving around regions with high likelihood, the user can easily obtain beautiful recolorized photos with the dedicated style of a particular artist. 

Another application of palette interpolation is photo recolorization with multiple local palettes. Conventional recolorization techniques only apply a single set of colors to an entire image via hue adjustment, which leads to unnatural recolorization.  Since an object does not simply contain the same color with different intensities, manual work is required to add more variations to these regions. For example, a tree is generally green but it can also contain blue-like or yellow-like variations of green. By taking the input colors and diversifying them with variations sampled from the palette density, we were able to produce results with more realistic colors (see Section~\ref{sec:photo_recolorization}).
Additionally, we developed an adaptive color palette that changes with the content being painted. The idea is, since selecting colors is time consuming and requires expertise in color mixing, it would be useful to have a palette that automatically presents to the user with compatible color themes, which fit well to the current content (Figure~\ref{fig:teaser}). For example, as a tree is being painted, the palette will suggest different variations of green (for the leaves) and variations of brown (for the branches). All of the mentioned applications are shown in the video accompanied with this work.

Finally, we carried out quantitative evaluation to measure the performance of our approach with relevant techniques. We conducted an experiment on the task of predicting the missing colors of a palette using GPLVM and other relevant methods. More specifically, given a number of observed colors (e.g., 4, 3, 2, or 1) in the palettes, we predicted the remaining colors of a 5-color palette. The experimental results show that our method yields satisfactory accuracy and is significantly better than previous methods such as ~\cite{o2011color,lin2013probabilistic}. 
\paragraph*{\textbf{Terminology}} The term ``palette''  used in our work is the same as ``color theme'' that appears in previous works.  We use both terms interchangeably.
 \vspace{-5pt}
\section{Related Work}\label{sec:related_works}
In this section, we discuss major research fields related to the color and theme studies.
\vspace{-10pt}
\subsection{Color Harmony}
Color is a popular research topic in many fields of arts and sciences. In psychology, for instance, various models have been proposed to capture the harmony of color combinations~\cite{schloss2011aesthetic,ou2006colour}.
These models often aim at deriving common ``laws'' of color compatibility by conducting psychological experiments on human perception. On the other hand, the graphics community is more interested in modeling color harmony for design and visualization. Cohen Or et al.~\cite{cohen2006color} directly apply harmony templates derived from classic color theories to the task of color harmonization.  Nishiyama et al.~\cite{Nishiyama2011} make use of color harmony to assess the quality of photographs.  Sauvaget et al.~\cite{sauvaget2010segmented} use color harmony for segmented photo colorization.  O'Donovan et al.~\cite{o2011color} take a data-driven approach by learning to predict color ratings from large datasets. They show interesting applications such as color suggestion, theme extraction, and theme optimization. Later the same authors propose a collaborative filtering approach to capture user preference of color combinations~\cite{o2014collaborative}. Recently, Lin et al.~\cite{lin2013probabilistic} incorporate the rating regression model in a factor graph that assigns harmonious colors to 2D patterns. Similar to our method, their work also performs palette interpolation but relies on a computationally expensive inference technique \---\ the Markov chain Monte Carlo method  (MCMC) \---\ which is unsuitable for real-time applications. Besides, the function that actually deals with input data only considers pair-wise relationships between colors in the palettes, while the global compatibility function is not trained on the input data. In contrast, our compatibility prediction comes from the probability density function, which is trained on the input data.
 In our color prediction experiment, we show that our method produces more accurate prediction than the methods introduced in~\cite{o2011color,lin2013probabilistic}. Moreover, our method is also fast enough to be integrated into real-time applications such as photo-style explorer and real-time color suggestion (see Section~\ref{sec:apps}).
\vspace{-10pt}
\subsection{Image Recolorization and Stylization}
Our work is also related to the problem of image recolorization and stylization. Color transfer is a standard approach for image colorization \cite{reinhard2001color}. The basic idea is to map the color distribution from a reference image to a given image. To model the color distribution, one can first extract color themes from reference images and then perform transferring~\cite{bonneel2013example,reinhard2001color,wang2010data}. Alternatively, histograms can be used to directly map the colors between images \cite{Neumann:2005}.  The method of Liu et al.~\cite{CGF:CGF12409} automatically discovers styles in a collection of user images via a clustering approach. The major difference between this work and ours to image recolorization is that, instead of clustering the photo collections, we applied \emph{manifold learning}~\cite{ma2011manifold} on palette data to interpolate color palette on demand. The interpolation allows us to generate variations of the existing color themes, thus providing the user with a wider range of color choices. 

Chang et al.~\cite{chang2015palette} provide an intuitive interface to recolorize an image by manually modifying a global color palette representing the image. Our work also introduces a method to recolorize an image with palettes. However, instead of letting the user modify the palettes, we automatically generate them from the palette manifold. 
Since our focus is to generate reasonable palettes rather than the recolorization technique itself, we do not provide comparative study with conventional methods in the field.  Chia et al.~\cite{chia2011semantic} introduce an interesting approach to grey-scale image colorization by retrieving semantically similar photographs from the Internet. Nguyen et al.~\cite{nguyen2015data} also retrieve images of an object and model a color manifold for that object.  Note that our method models the manifold of \textit{color palettes} rather than \textit{colors}, which is more complicated since we need correspondences between colors. Consequently, our method is specifically designed for the purpose of color suggestion and color compatibility prediction.
\vspace{-10pt}
\subsection{Color Picking and Recommendation}
Color picking is an essential component of all design software. 
Basic color pickers provide a wheel-like interface, where the user drags the pointer around the gamuts to find the right color, or a slider-like interface that allows the user to adjust color components individually. More advanced tools, like Adobe Color CC~\cite{adobecolor2015} or Coolorus~\cite{coolorus2015}, help the user to choose harmonious color combinations by using harmony templates, which are derived from classic color theories~\cite{itten1970elements,matsuda1995color,tokumaru2002color}. Another strategy is to clip the color gamut to limit the color choices to ``harmonious groups''~\cite{coolorus2015}. For people who are acquainted with physical palettes, Corel Painter~\cite{corelpainter2015} and MyPaint~\cite{mypaint2015} provide color mixer pads that simulate real-world color mixing. Gradient mixer~\cite{meier2004interactive} is another tool that displays gradients between colors to add more variations to the current palette. Our model allows interpolation between palettes, thus, at some points on the manifolds, one can find ``in-between'' colors similar to  the gradient mixer. Palette breeder~\cite{meier2004interactive} is an interesting tool that uses a genetic algorithm to ``breed'' palettes and generates new ones. Wijffelaars et al.~\cite{wijffelaars2008generating} suggest an intuitive interface that uses curves to represent color harmony parameters. Using color names is another way to select color~\cite{heer2012color}. 

Nguyen et al.~\cite{nguyen2015data} use a color manifold and apply dimensionality reduction to ease the task of changing colors of an object. Our user interface also provides a mechanism for the user to continuously surf through variations of colorizations. As we also model the density of color themes, our interface assists users by displaying the probabilities associated with the variations, assisting them in browsing the theme space. The work of Shapira et al.~\cite{shapira2009image} suggests a promising parametric interface for browsing possible appearance of an image but does not provide recommendations to the user and is not data-driven. Similarly, Marks et al.~\cite{marks1997design} introduce a framework called Design Galleries that displays variations of a parametric graphical object in a reasonable way (e.g. images, animations...). Their method is not data-driven and their focus is to present an object's variations rather than to recommend variations, as in our work.
\vspace{-10pt}
\subsection{Object Arrangement and Browsing}
Our palette arrangement is close to the more generic problem of object arrangement. The objective is to arrange a set $A=\{a_1,...,a_N\} \in X $ of objects into a set $B=\{b_1, ...,b_N\} \in Y$ of locations so that the pairwise distances between objects in $A$ are preserved  as much as possible after being aligned into $B$. A popular solution for this problem is the ``kernelized sorting'' technique introduced in ~\cite{quadrianto2009kernelized}, in which the authors utilize kernels and a learned centering matrix to make distances in $X$ and $Y$ comparable. More recently, Fried et al. elaborate a method to match a set of objects with a layout called IsoMatch~\cite{fried2015isomatch}. Although our method is similar to IsoMatch in a spirit, the goals of the two methods are different. While IsoMatch aims at arranging N objects to N locations, our problem is to solve a set of M such problems, which are interdependent.

\section{Methodology}
As illustrated in Figure~\ref{fig:teaser}, our method for modeling the density of color palettes can be divided into 3 steps: 
\begin{itemize}
\item \textbf{Data Collection and Preprocessing}: We collected images of different artists and extracted color themes from local regions.
\item \textbf{Palette Ordering}: We ordered the colors in each palette in a consistent way.
\item \textbf{Learning and Inference}: We applied density estimation techniques and performed interpolation on palette data.
\end{itemize}
\vspace{-10pt}
\subsection{Data Collection and Preprocessing}\label{sec:dataset}
\begin{figure}[ht]
  \centering
  \includegraphics[width=0.8\linewidth]{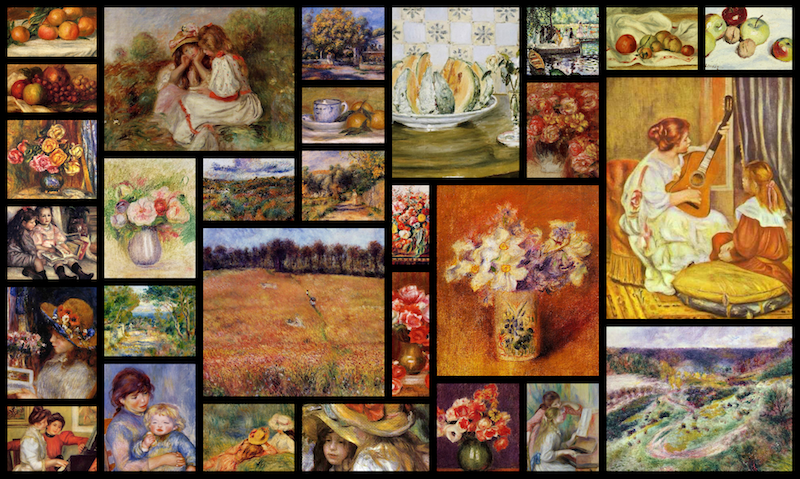}
  \includegraphics[width=0.8\linewidth]{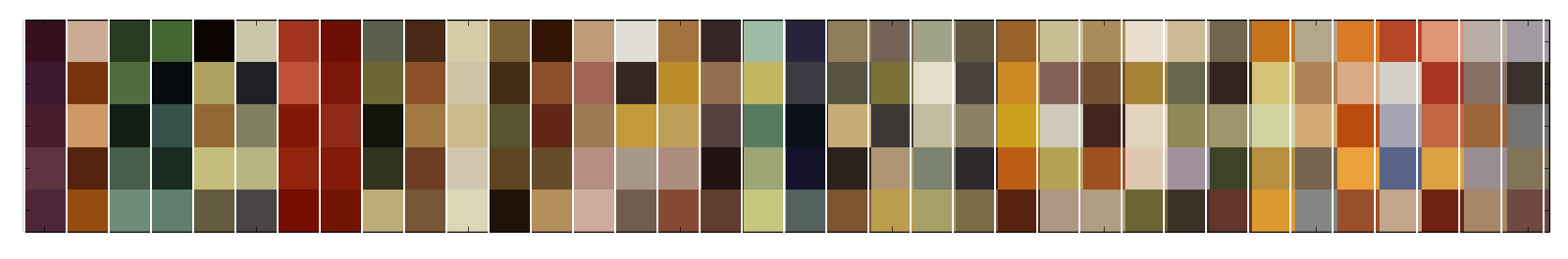}
  \includegraphics[width=0.8\linewidth]{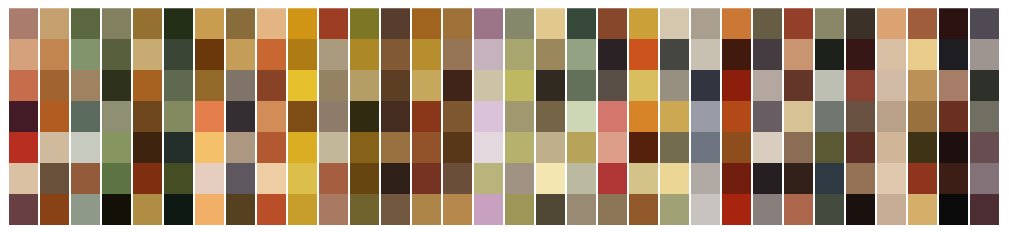}
  \caption{Our dataset contains a set of 40 images from each of 17 well-known artists. (Top) paintings from Renoir set. (Mid) 5-color palettes extracted from the set. (Bottom) 7-color palettes extracted from the set. Each column is a color palette.}\label{smartpalette:fig:dataset}
\vspace{-10pt}
\end{figure}

As partially shown in Figure~\ref{smartpalette:fig:dataset} (Top), we collected 40 paintings for each of 17 well-known artists such as Pierre-Auguste Renoir (Renoir), Vincent Willem van Gogh, Raffaello Sanzio da Urbino (Raphael), Maximilien Luce (Luce), Paul C\'{e}zanne (Cezanne), Hilaire-Germain-Edgar De Gas (De Gas), and William Merritt Chase (Chase). All the paintings were rescaled so that the larger dimension is 500 pixels. As explained in Section~\ref{smartpalette:sec:intro}, instead of treating a whole painting as a single palette,  we extracted palettes from local patches from the paintings, which capture the local contexts. Specifically, we extracted patches of 200x200 pixels from the paintings using a sliding window with a step-size of 100 pixels. 
Next, we selected 1000 pixels from each patch by randomly sampling from the patch. Subsequently, we applied K-Means in the CIE*Lab space to cluster the pixels in each patch to obtain $K$ clusters (Figure \ref{smartpalette:fig:dataset} (Mid)), thus forming one color palette per patch.  We randomly selected 400 palettes per artist and aggregated them together to form a dataset of color palettes \---\ one dataset per artist. We call our palette datasets the \emph{Artistic Palette Datasets}. For quantitative study, we chose $K=5$, which follows previous works~\cite{o2011color,lin2013probabilistic}. Our methodology is, however, independent from the size of the palette. Thus, the number of colors can be chosen arbitrarily, depending on the application. For some applications, such as photo recolorization and multi-style painting, which often need more colors, we used $K=7$ (Figure \ref{smartpalette:fig:dataset} (Bottom)). We also used $K=10$ for demonstration purpose in Section ~\ref{sec:evals}. Please refer to Section~\ref{smartpalette:sec:eval_ordering} for more discussions about $K$.
\subsection{Palette Ordering}\label{sec:reordering}
In this section, we discuss the palette ordering problem and our solution. We also show comparisons with some other methods in Section~\ref{smartpalette:sec:evalsort}.
\subsubsection{Problem Statement}
\begin{figure}[ht]
\centering
\includegraphics[width=0.8\linewidth]{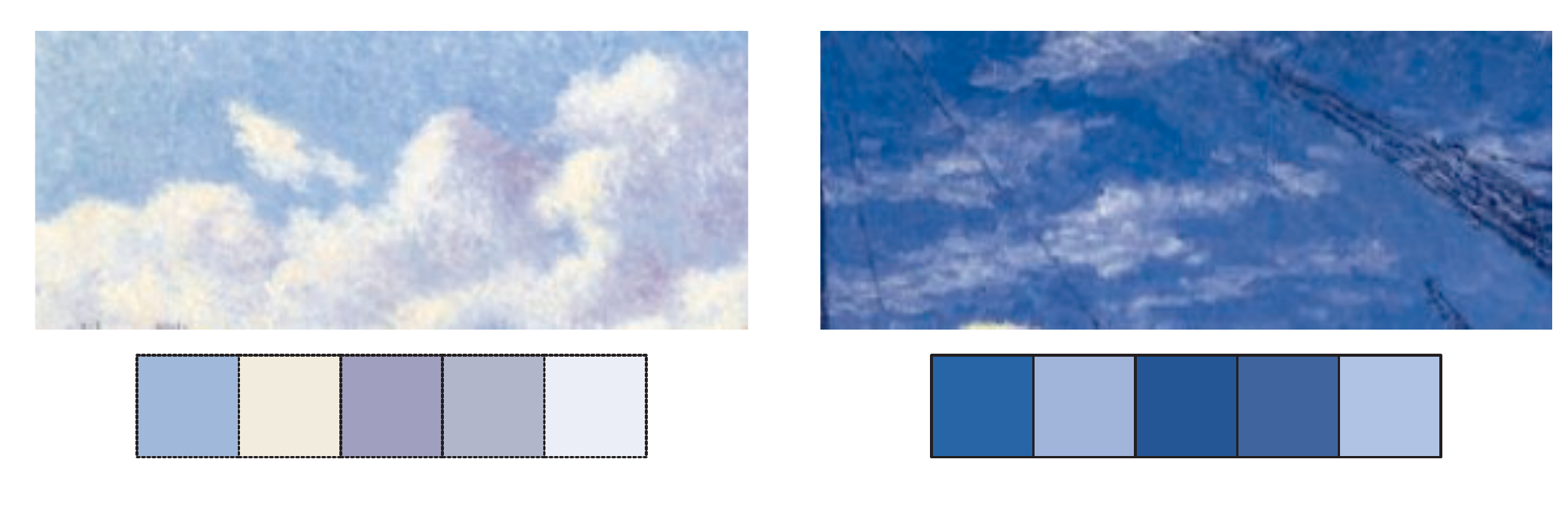}
\caption{Palettes extracted from similar patches often have correspondences. The extracted palettes were ordered with Binary Palette Sort.}\label{smartpalette:fig:correspondence}
\vspace{-15pt}
\end{figure}
The goal of palette ordering is to order the colors in each palette, so that linear interpolation between ordered palettes yields meaningful results \---\ i.e., the colors that describe the same objects should have the same indices. The intuition is, the palette dataset might contain palettes that come from the same context, and thus there are always semantic correspondences between the colors. Figure \ref{smartpalette:fig:correspondence} shows an example, in which both patches contain sky scene and one can observe the correspondences between cloud colors and between sky colors. It is, however, too costly to manually annotate all semantic regions as an image might contain hundreds of regions. We overcome this difficulty by assuming that patches containing similar sets of colors should have similar contexts. Our algorithm can guarantee that similar palettes are optimally aligned. Very different palettes are only roughly aligned via palette set ordering, as described in Section~\ref{smartpalette:sec:binarysortalgo}.

Formally, given a palette dataset $X=\{\mathbf{P}_1, ... , \mathbf{P}_N\}$, with $N$ is the number of palettes. We have the palette $\mathbf{P}_n = \left[ \mathbf{p}_n^1, \hdots,  \mathbf{p}_n^K \right]^T $ , where $\mathbf{p}_n^k$ is a color (a 3D vector).  
The goal is to solve the following optimization problem, which minimizes the pairwise distances between the palettes:
\begin{equation}\label{smartpalette:eqn:globaopt}
\mathrm{argmin}_{\mathbf{g}} \sum_{n=1}^{N} \sum_{m=1}^{N} \sum_{k=1}^{K} \lVert \mathbf{p}_{n}^{g_n(k) } - \mathbf{p}_{m}^{g_m(k) } \rVert, 
\end{equation}
 where $g_n: \mathbb{K} \rightarrow \mathbb{K},~\mathbb{K} = \{ 1, \hdots, K \} $ is a bijection that maps an index $k$ to another index $k'$. Solving the palette ordering problem thus has the complexity $\mathcal{O}( (K!)^N )$, which grows exponentially with the number of palettes.
 
Given that each dataset in our Artistic Palette Datasets contains 400 hundred palettes, it is not feasible to solve the problem ~(\ref{smartpalette:eqn:globaopt}) with brute-force. To this end, we introduce the Binary Palette Sort algorithm (BPS) \---\ an ordering algorithm that uses a divide-and-conquer strategy to achieve a reasonable solution for~(\ref{smartpalette:eqn:globaopt}). 
 \begin{figure}[ht]
\centering
\includegraphics[width=1.0\linewidth]{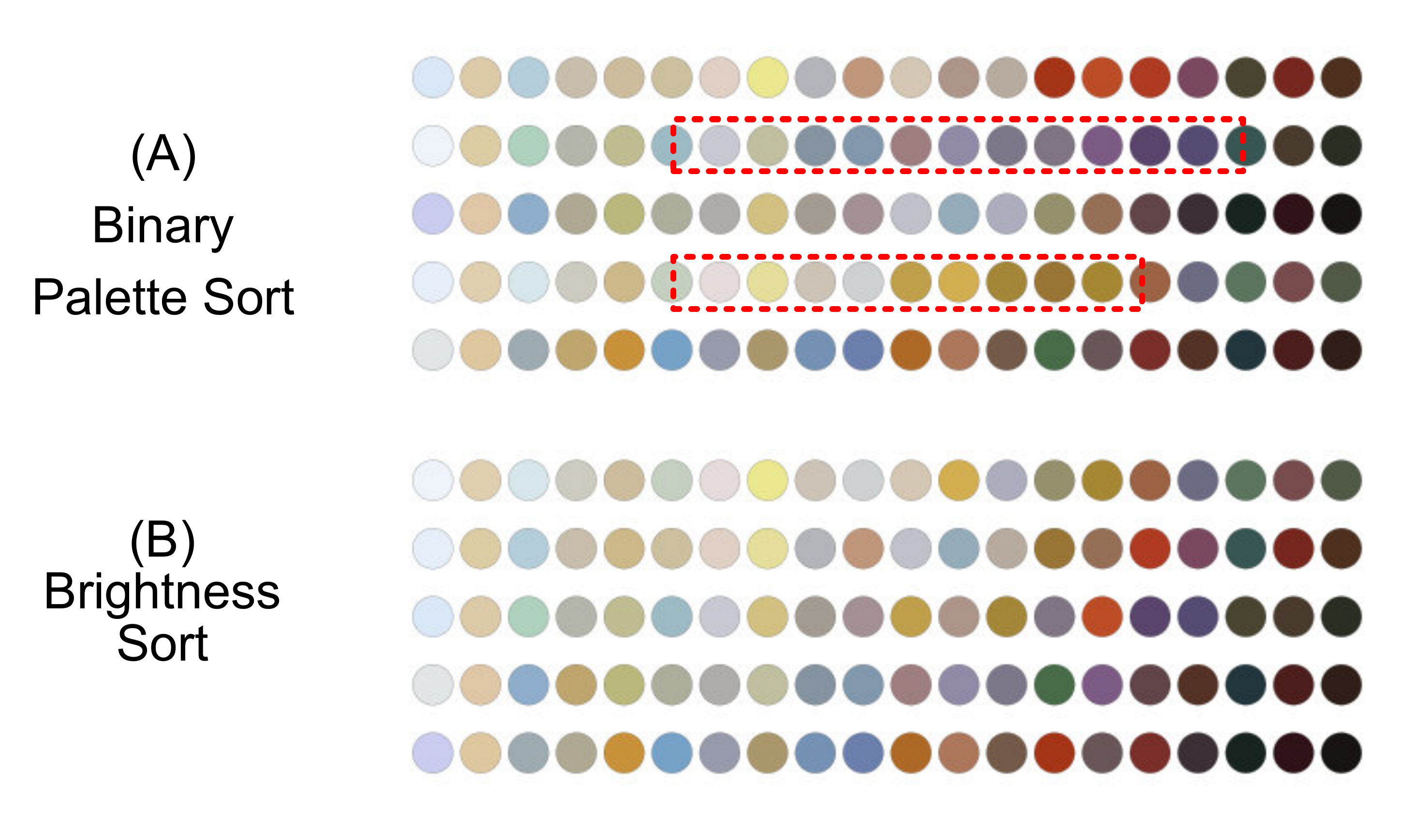}
\caption{\huy{Example of palette ordering with our method. (A) Binary Palette Sort and (B) Unordered. Each column is a paletete. The red dot boxes in (A) highlight the rows in which colors are smoothly arranged.}}\label{fig:comparesort}
\end{figure}

We motivate the strategy of BPS by showing a typical palette dataset in Figure~\ref{fig:comparesort} (B). As a palette dataset might contain color palettes that are very different from each other, it is less important to align such palettes, comparing to the similar ones. For instance, the first palette (first column) in Figure~\ref{fig:comparesort} (B) represents a sky and thus aligning it to the last palette (last column), which comes from a tree, does not make sense. This is the key observation that motivates our sorting algorithm. We prioritize the matching between nearby palettes over distant palettes. Distant palettes are still roughly aligned via set alignment, which will be discussed later.
 
  \begin{figure}[ht]
\centering
\includegraphics[width=0.95\linewidth]{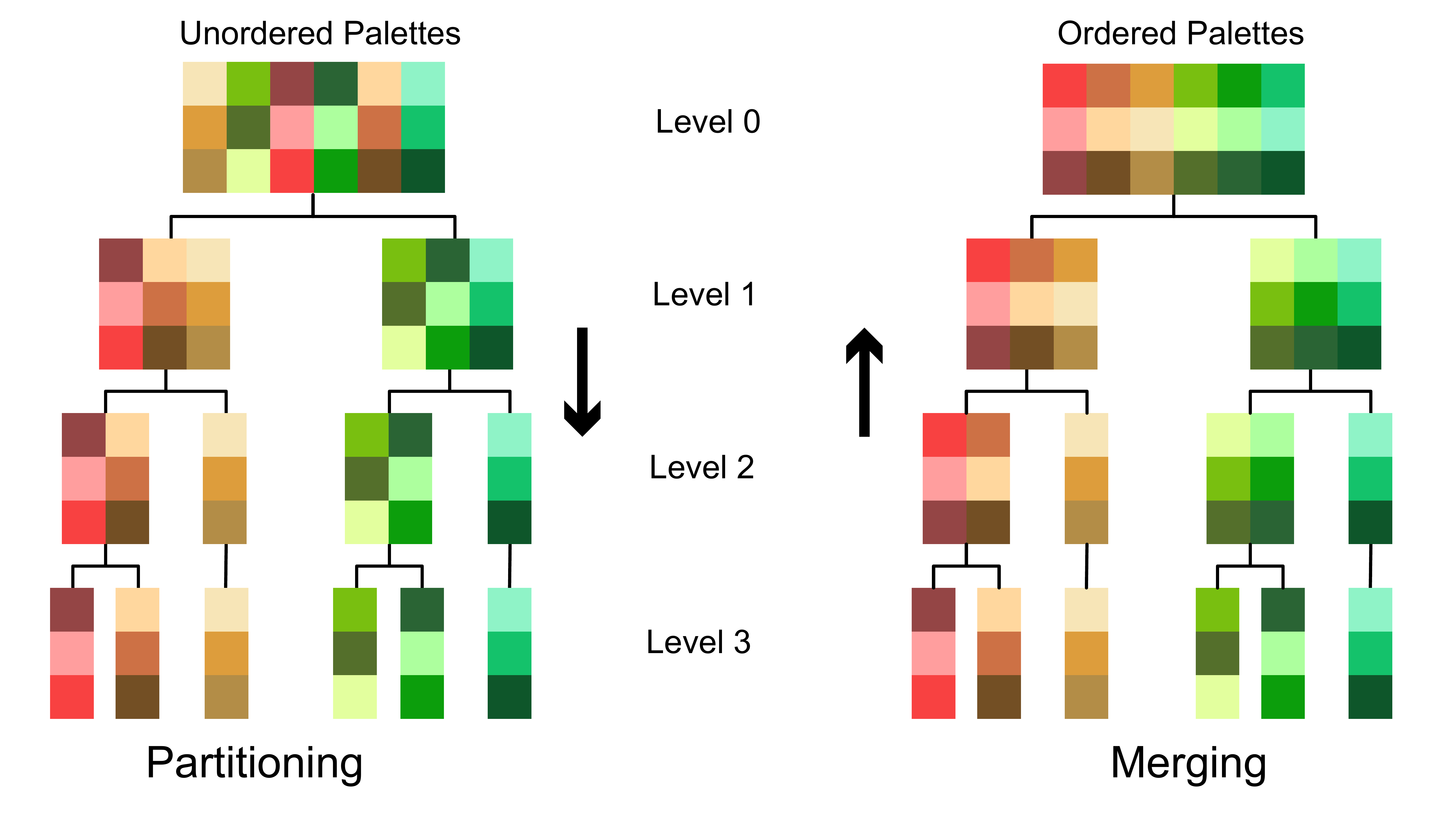}
\vspace{-10pt}
\caption{Example of palette arrangement with Binary Palette Sort. The unordered palettes were partitioned by considering the similarities between palettes.}\label{fig:binarysort}
\vspace{-15pt}
\end{figure}
\subsubsection{The Binary Palette Sort Algorithm}\label{smartpalette:sec:binarysortalgo}
Intuitively, BPS recursively partitions the dataset $X$ into two subsets until there is only one palette left in each subset and then works the way back to the first call by aligning and merging each pair of subsets into an ordered list. The result is a set of sorted palettes (see Figure \ref{fig:binarysort}).  Our BPS algorithm is similar to Merge Sort~\cite{thomas2001introduction} in strategy but has a different merging method and a different objective. In our context, the objective of Merge Sort would be to rearrange a list of color palettes while the objective of Binary Palette Sort is to rearrange the colors in each palette, while the order of the palettes is fixed. BPS has a complexity  of $\mathcal{O}(N K^4)$, which is feasible to solve. At each level of recursion, we solve a sorting sub-problem that aligns two sets of color palettes, where each set can have a different number of palettes (see the description of \emph{Sort} below).

Given the palette dataset $X$, our algorithm to approximate the solution for~(\ref{smartpalette:eqn:globaopt}) is expressed recursively as
\begin{multline}
\mathrm{Sort}(X) = \mathrm{Merge}( \mathrm{Sort}(\mathrm{Partition}_{l}(X)), \\
\mathrm{Sort}(\mathrm{Partition}_{r}(X)) ),
\end{multline}
where we use subscripts $l$ and $r$ to denote the two partitions produced by the subroutine $Partition$. Each subroutine is given in details below.
\paragraph*{\textbf{Sort}} The sort subroutine takes a palette dataset $X$ and returns a new palette dataset $X'$, whose each member palette is aligned.  The routine starts from the lowest level of the tree, solving a case of \emph{linear assignment problem} (LAP), in which both $P'$ and $Q'$ are singletons, merging the aligned sets together and moving up to the higher level. If a leaf node does not have a ``partner'' to match (the rightmost leaf nodes in Figure \ref{fig:binarysort}), the routine just passes it to the upper level. At the upper level, the subroutine repeats the assigning and merging processes that have been done at the lower level. Note that now each set may contain a different number of palettes (see Figure \ref{fig:binarysort} (Right)). 

To align two sets of palettes, we need to solve a linear assignment problem as follows:  Given two sets of color palettes $P =\{\mathbf{P}_n \}_{n=1}^{N}$ and $Q =\{\mathbf{Q}_n\}_{n=1}^{M}$, where $N, M$ are the numbers of palettes, we have the color palettes $\mathbf{P}_n=\left[\mathbf{p}_n^1, \hdots, \mathbf{p}_n^K \right]^T, \mathbf{Q}_n=\left[\mathbf{q}_n^1, \hdots, \mathbf{q}_n^K \right]^T$, where $K$ is the number of colors in each palette and $\mathbf{q}_n^k, \mathbf{p}_n^k$ are the colors. 

By grouping corresponding colors in the palettes in each set together, we obtain matrices $\mathbf{P}^k = \left[ \mathbf{p}_1^k, \hdots, \mathbf{p}_N^k  \right]^T$ and $\mathbf{Q}^k = \left[ \mathbf{q}_1^k, \hdots, \mathbf{q}_M^k  \right]^T$ which can also be viewed as color palettes. In Figure~\ref{fig:binarysort}, $\mathbf{P}_n$ and $\mathbf{Q}_n$ are the matrices (blocks of colors), each column is a palette. $\mathbf{P}^k$ is the k-th row in the color matrix, which will form the corresponding colors. We want to match the rows of $\mathbf{P}_n$ and $\mathbf{Q}_n$ so that they are aligned well according to the distance $d$ (see below). The problem is stated as:
\begin{equation}\label{smartpalette:eqn:newopt}
\mathrm{argmin}_{g}\sum_{k=1}^K f(\mathbf{P}^k, \mathbf{Q}^{g(k)} ),
\end{equation} where $f$ is the distance between sets of colors and $g$ is similar to that of (\ref{smartpalette:eqn:globaopt}). In our implementation, $g$ is found with the Hungarian algorithm~\cite{kuhn1955hungarian} and $f$ is given below.
 Since the palettes $\mathbf{P}^k$ and $\mathbf{Q}^k$ have different lengths, we cannot apply conventional metrics for vectors such as the Euclidean distance. Instead, we utilize the modified Hausdorff distance (MHD) introduced in~\cite{dubuisson1994modified}. The MHD between two palettes of different lengths $\mathbf{P}^k$ and $\mathbf{Q}^h, ~\text{for } h=g(k)$ is calculated as follows:
\[ \max(d(\mathbf{P}^k, \mathbf{Q}^h), d(\mathbf{Q}^k, \mathbf{P}^h))
, \]
where 
\[
d(\mathbf{P}^k , \mathbf{Q}^h) = \frac{1}{N} \sum_{i=1}^{N} \min_{j \in 1:M} \lVert \mathbf{p}_i^k - \mathbf{q}_j^h \rVert.
\] $d(\mathbf{Q}^h, \mathbf{P}^k)$ can be defined similarly. 
The process is repeated until all the palettes in the dataset are merged into a single set. At this point, we have already obtained a sorted set of  palettes where the colors in the same row are close to each other. 
\paragraph*{\textbf{Merge}}
The $Merge$ subroutine merges two sets of color palettes into a new set, in which the palettes are aligned with the indices $g(k)$ is taken from (\ref{smartpalette:eqn:newopt}). Figure~\ref{fig:binarysort} (Right) gives an example of this step.
\begin{equation}\label{smartpalette:eqn:merge}
    \mathrm{Merge}(P,~Q) = 
    \begin{cases}
     	P ,~\text{for } Q = \emptyset \\
     	Q ,~\text{for } P = \emptyset \\
     	P \cup
     	\left\lbrace
     	\left[
		\begin{matrix}
		\mathbf{q}_1^{g(1)}      \\
		\vdots \\
		\mathbf{q}_1^{g(K)} 
		\end{matrix}
		\right]
		, \hdots ,
     	\left[
		\begin{matrix}
		\mathbf{q}_M^{g(1)}      \\
		\vdots \\
		\mathbf{q}_M^{g(K)} 
		\end{matrix}
		\right]
		\right\rbrace
   \end{cases}.
\end{equation}
\paragraph*{\textbf{Partition}}
This subroutine divides the set $X$ into two subsets $X_l, X_r, X_l \cap X_r = \emptyset $ while minimizing the total distance from the data points in each subset to the center of the subset. The goal is to ensure that palettes in each subset are close to each other.
\begin{equation}\label{smartpalette:eqn:partition}
\mathrm{argmin}_{\gamma, \mu_l, \mu_r} 
\sum_{n=1}^{N} \gamma(\mathbf{P}_n) \bigl \lVert   \mathbf{P}_n -  \mu_l \bigr  \rVert + 
\sum_{n=1}^{N} (1 - \gamma(\mathbf{P}_n)) \bigl \lVert   \mathbf{P}_n -  \mu_r   \bigr \rVert ,
\end{equation} where $\mu_l,~\mu_r$ are the centers of each subset and $\gamma$ is: 
\[
\gamma(\mathbf{P}_n) = 
\begin{cases}
0, \forall \mathbf{P}_n \not\in X_l \\
1, \forall \mathbf{P}_n \in X_l
\end{cases}.
\] 
We then have 
\begin{align}\label{smartpalette:eqn:partition_a}
X_l & = \{ \mathbf{P}_n | \forall \mathbf{P}_n \in X, \gamma(\mathbf{P}_n) = 1 \}, \nonumber \\
X_r & = \{ \mathbf{P}_n | \forall \mathbf{P}_n \in X, \gamma(\mathbf{P}_n) = 0 \}.
\end{align}
This process is repeated until each subset contains only a single palette, resulting in a tree in which each node is a set of palettes and the leaf nodes are singletons. Although there are solutions such as the K-medoids algorithm for the problem (\ref{smartpalette:eqn:partition}), it could be costly to apply them in practice. To speed up the algorithm, we present an alternative partition method that has comparable performance and is much faster. Intuitively, the new partition method projects the palettes onto a line and then orders the palettes according to the projected coordinates. The dataset is then partitioned by the palettes' indices. Specifically, we used kernel principal component analysis (KPCA)~\cite{Scholkopf:1999}, which can operate on sets, to project our palettes. The kernel used was derived from the MHD distance introduced above (MHD kernel). Although there are other powerful kernels like the pyramid match kernel (PMK)~\cite{grauman2005pyramid}, which work on sets, we found that the MHD-based kernel is sufficiently accurate and is the fastest method, allowing it to work in real-time. In fact, as PMK was designed for matching large sets, it did not perform well on our palette data and was also too slow for our applications. The left side of Figure \ref{fig:binarysort} demonstrates this process.

\begin{figure}[ht]
\centering
\includegraphics[width=0.8\linewidth]{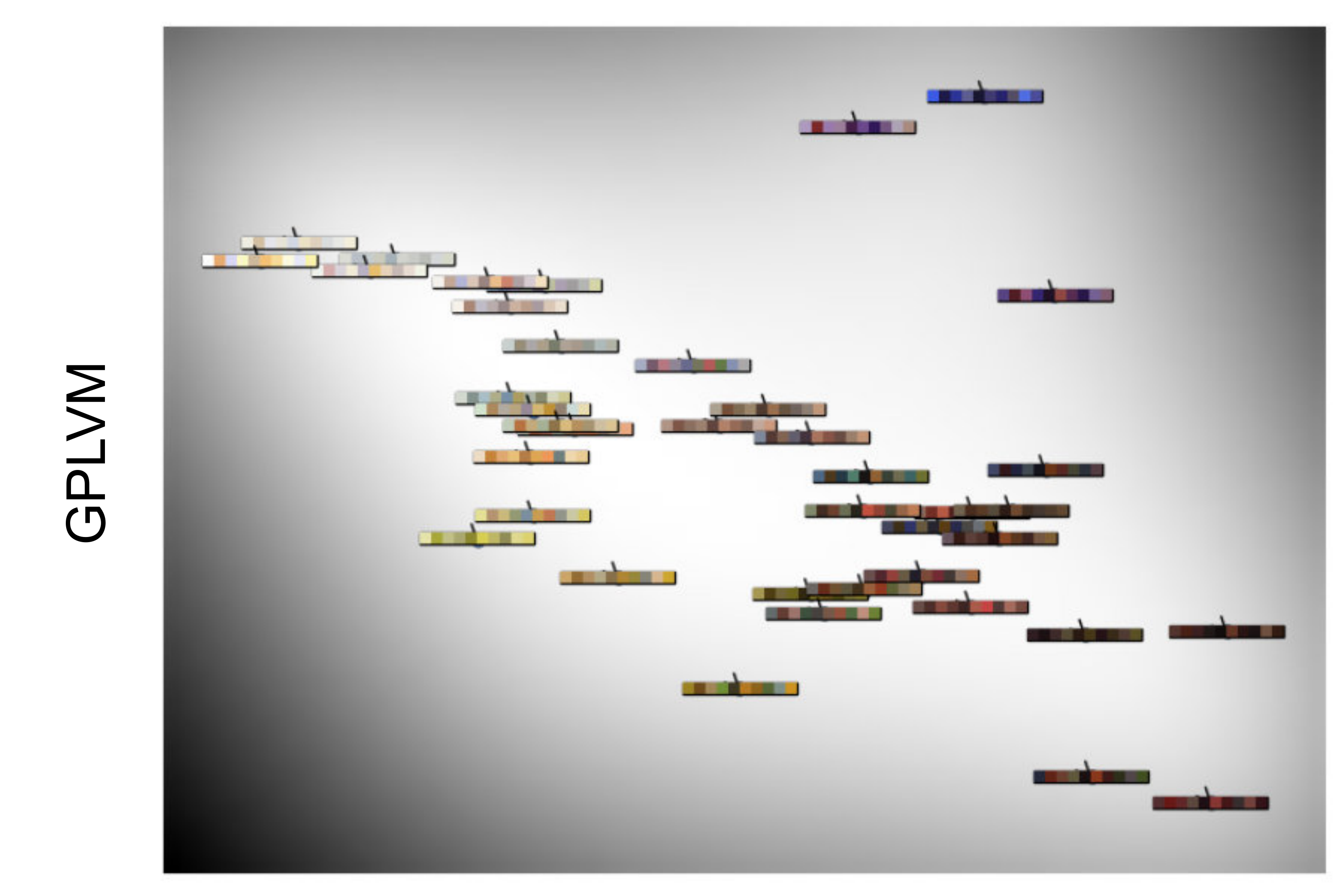} 
\vspace{-10pt}
\caption{A sample palette density produced by GPLVM. Gray values indicate the likelihoods of the palettes.}\label{fig:gmm_gplvm_density}
\vspace{-10pt}
\end{figure}
\subsection{Learning and Inference}\label{sec:density}
Prior to learning and inference, we directly concatenated colors in an ordered palette to form a vector of dimensionality 15 (for 5 colors), 21 (for 7 colors) and 30 (for 10 colors). Given the concatenated dataset, we performed \emph{palette interpolation}, \emph{density estimation}, and \emph{color prediction}. Although there are many possible approaches to our problems such as neural networks~\cite{hinton2006reducing} and latent Dirichlet allocation~\cite{blei2003latent}, we chose a standard method named Gaussian process latent variable models (GPLVM)~\cite{lawrence2004gaussian} for its ability to deal with small datasets, which is the case of our work.
\vspace{-5pt}
\subsubsection{Palette Interpolation}\label{smartpalette:sec:palette_interpolation}
GPLVM was first introduced by Lawrence et al. for the purpose of nonlinear dimensionality reduction (NLDR) and visualization~\cite{lawrence2004gaussian}. It shares the same objective function with popular dimensionality reduction methods such as KPCA, Probabilistic PCA, and Multidimensional Scaling (MDS)~\cite{kruskal1964multidimensional}.  The main advantage of GPLVM over other methods is the non-linear mapping from latent space to observed space. This means that it is suitable for non-linear data interpolation, which is the case of our work. We interpolated new palettes by sampling data points from the low dimensional space induced by GPLVM. GPLVM provides a mechanism to project these latent points back to the palette space via Gaussian Processes (GPs). The main computation for training a GPLVM is the nonlinear minimization of the negative log-likelihood function with respect to the kernel parameters and the latent points~\cite{lawrence2004gaussian}. This minimization was done with a gradient-based optimizer, the scaled conjugate gradient method (SCG).
\paragraph*{\textbf{Manifold learning}}
A concept related to NLDR is \emph{manifold learning}~\cite{ma2011manifold}, one common approach to NLDR. Manifold learning aims to find an embedded non-linear topological manifold in a high dimensional space. By projecting data onto a low dimensional manifold, one can visualize complex data in an understandable manner. In our work, we are interested in the inverse projection that maps the low dimensional data points back to the original space. The idea is to let the user interact with palette data in a low dimensional space \---\ 2D, in our case \---\ and to observe the results of these interaction in the palette space.
\vspace{-5pt}
\subsubsection{Density Estimation}\label{smartpalette:sec:density_estimation}
GPLVM assigns likelihoods to data points in the latent space, which can be interpreted as a \textit{density}. In our implementation, GPLVM was trained with the Radial Basis Function (RBF) kernel~\cite{vert2004primer}, and the number of latent dimensions was 4.  In Figure \ref{fig:gmm_gplvm_density}, we show an example of palette density, induced by GPLVM. The two most significant dimensions were chosen for visualization. The palettes were taken from 4 datasets: De Gas, Luce, Gogh and Raphael. For demonstration purpose, we chose the number of colors to be 10.

To justify our choice of the learning model, we also experimented with principal component analysis (PCA) for dimensionality reduction and Gaussian mixture models (GMM) for density estimation.The experiments show that GPLVM, as a non-linear method can produce more diverse results than PCA-GMM, which is a desired characteristic in our applications (see sample results for both methods in Section~\ref{smartpalette:sec:eval_interpolation}). In addition, the quantitative experiment in Section \ref{sec:evals} shows that GPLVM is slightly more accurate in predicting colors.

\vspace{-5pt}
\subsubsection{Predicting Compatible Colors}\label{sec:palette_completion}

To recover the missing colors with GPLVM, we first projected the partially observed palette $\mathbf{y}_*^o$ onto the latent space using the technique described in \cite{lawrence2004gaussian}. This technique involves the selection of an initial position for the latent point $\mathbf{x}_*$ by finding a similar point in the input space (palette space). The latent point is found by minimizing the same objective function as in the training step with respect to $\mathbf{x}_*$ while $\mathbf{y}_*^o$ and other parameters are fixed. GPLVM handles the missing dimensions by setting the corresponding noise variances to infinity. The likeliness between the predicted palette and the query palette can be controlled by setting the number of iterations in the minimization step. To recover the palette, we used a standard GPs prediction that maps data points from the latent space to input space. Both of the mentioned methods (PCA-GMM and GPLVM) are fast and can be used for real-time prediction.
 \vspace{-10pt}
\section{Applications}\label{sec:apps}
In this section, we introduce various applications that were based on our palette interpolation and completion methods. 


\subsection{Photo Recolorization}\label{sec:photo_recolorization}
Our method can suggest colors for photo recolorization. Different from traditional color transfer methods, which copy lightness, hue, and saturation from one photo to another, our method summarizes color characteristics of a set of source photos (image patches), and then actively examines regions in the destination photo to infers a \textit{set of colors} for each region. This makes our method different from that of \cite{huang2014learning,wang2010data}, which only give a single color to an image region (segment). The advantage of this approach is that local shades are preserved instead of being erased like in the cases of \cite{wang2010data,huang2014learning}. More details about our recolorization method is provided in the supplementary.

Figure \ref{fig:colorization} (Top) shows examples of photo recolorization with color styles learned from different artists. Observe how the vivid blues in Luce set and the strong oranges in Cezanne set are well captured by our method. For the case of Chase set, one might notice the dominance of fire/canary yellows and crimson reds. The results from Raphael set are less vibrant and bear the classic style of the artist. \huy{Additionally, we show the results produced with the Beach set in Fig. \ref{fig:context_specific}. Notice how the colors of sand, sunset sky and blue sky have been transferred to objects in the source images.}

\begin{figure}
\centering
\includegraphics[width=1.0\linewidth]{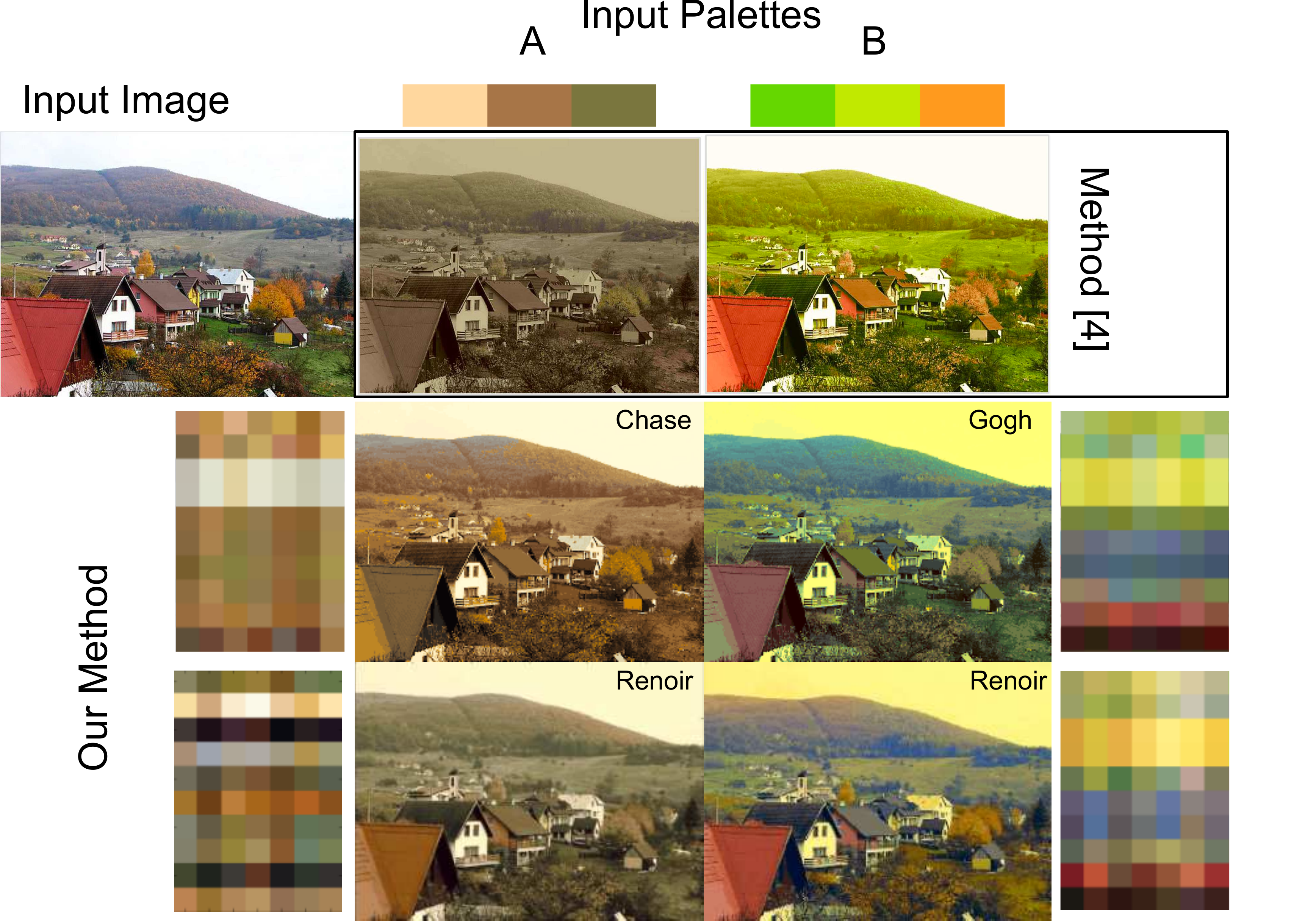} 
\caption[SmartPalette: Comparing recolorization results.]{Recolorization results obtained by our method and the method in \cite{wang2010data}. (A) an (B) are the user input palettes. The palettes predicted by our method are shown next to each result.}\label{smarpalette:fig:compare_colorization}
\vspace{-10pt}
\end{figure}


To see how recommending multiple palettes instead of one global palette would affect visual quality of the results, we compared our recolorization method with the method introduced in \cite{wang2010data}. Figure ~\ref{smarpalette:fig:compare_colorization} shows our recolorization results given the same input image and the target themes (A) and (B). To perform recolorization with an input palette as the constraint, we first applied the input theme to each segment in the destination image and then used the same steps as in the standard method.
We used the models trained separately on 3 sets to suggest colors: Chase, Gogh and Renoir. One might notice that for the input palette (A), our method produce much cleaner results, as compared with the result from \cite{wang2010data}, and still resembling the spirit of the input. At small scales, as our method suggests multiple colors for a single region \---\ the mountain, for instance \---\ we can observe a wider range of colors exists in each region, creating a richer look for the image. We call this approach to photo recolorization \emph{enriched photo recolorization}. Similar results can be seen in the case of the input (B). Note that, as the artists rarely used such vibrant colors in their paintings, images in (B) are less saturated than the input palette. Depending on the application, one can prepare a more appropriate dataset to achieve desirable results.
 
\begin{figure*}[ht]
\centering
\includegraphics[width=0.8\linewidth]{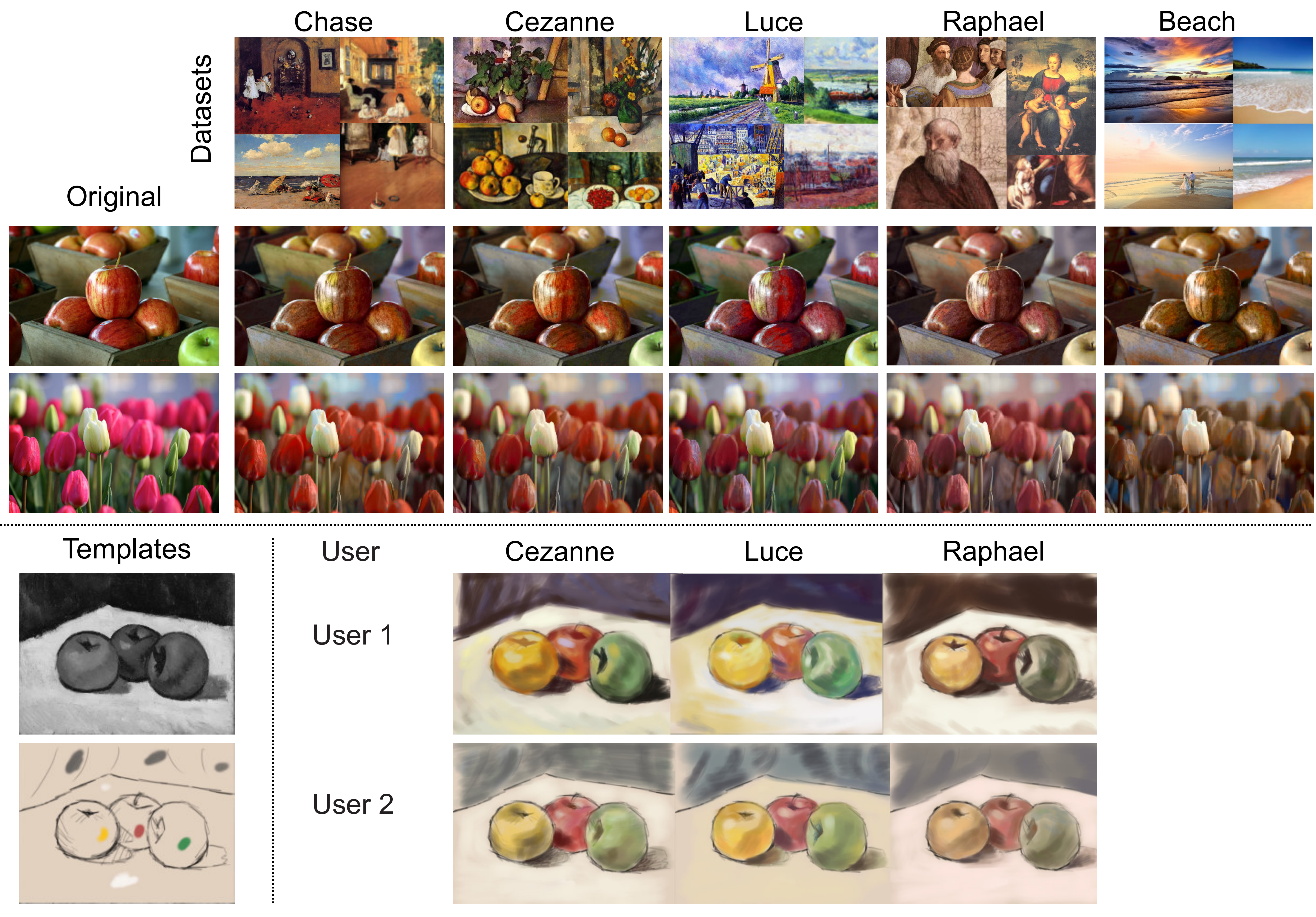}
\vspace{-10pt}
\caption[SmartPalette: Image recolorization and multi-style painting results.]{Image recolorization and multi-style painting results. (Top) Recolored images with palettes suggested by models trained on Chase, Cezanne, Luce, Raphael and Beach sets. (Bottom) Paintings made by two users using colors suggested by 3 models (Cezanne, Luce and Renoir). A pair of images were used as guidance (template). The grayscale image was shown to help the users with object shading and the line sketch with color spots was used to indicate the object boundaries and to guide the users on which color to use.}\label{fig:colorization}
\vspace{-10pt}
\end{figure*}
\begin{figure}[ht]
\centering
\includegraphics[width=0.85\linewidth]{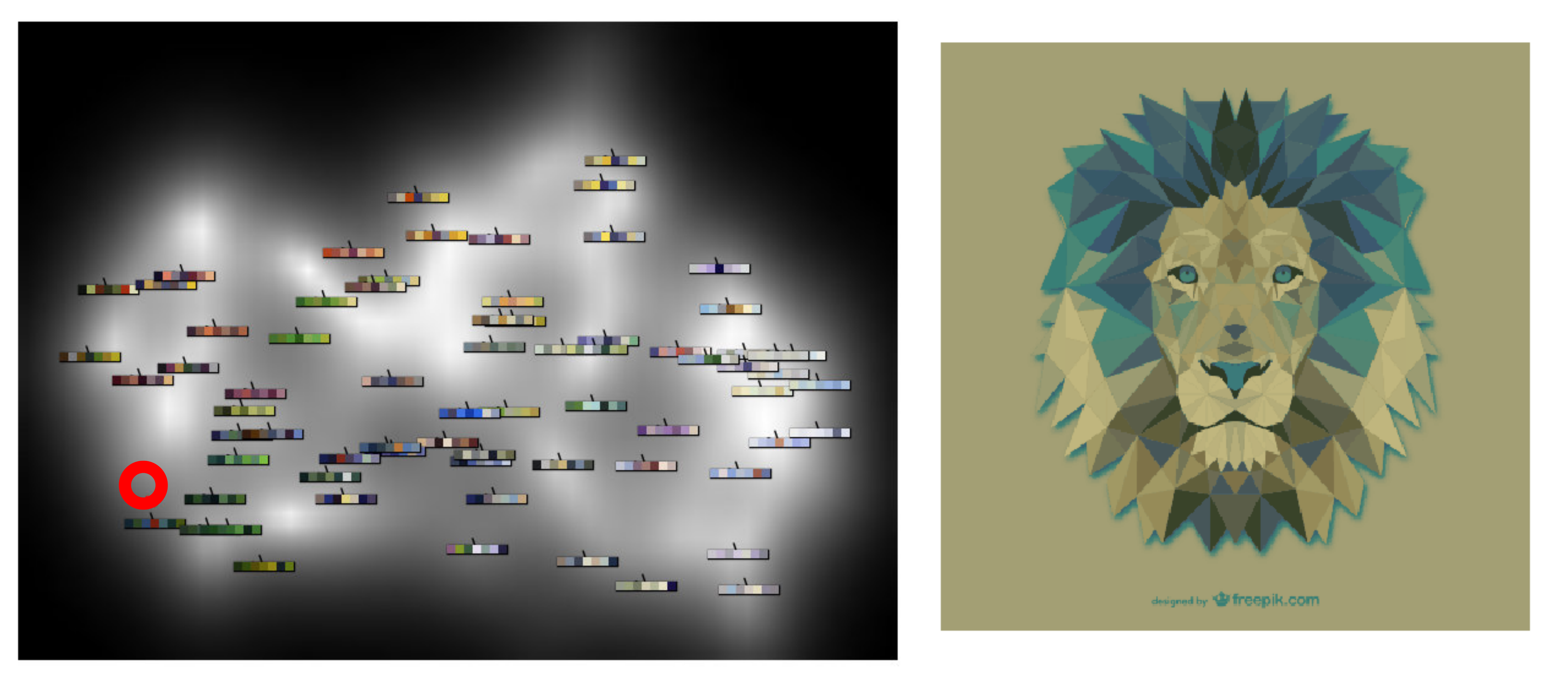}
\caption[Photo-style Explorer interface.]{Photo-style Explorer interface.  (Left) the space of color palettes where each point corresponds to a complete 7-color palette. Key points are annotated with the associated palettes and the red circle indicates the location of the mouse pointer. (Right) Image viewer, displaying the recolorized result in real-time.}\label{fig:explorer_interface}
\vspace{-10pt}
\end{figure}
\vspace{-5pt}
\subsection{Photo-style Explorer}\label{smartpalette:sec:photoexplorer}
\begin{figure*}[ht]
\centering
\begin{tabular}{cc}
\includegraphics[width=0.50\linewidth]{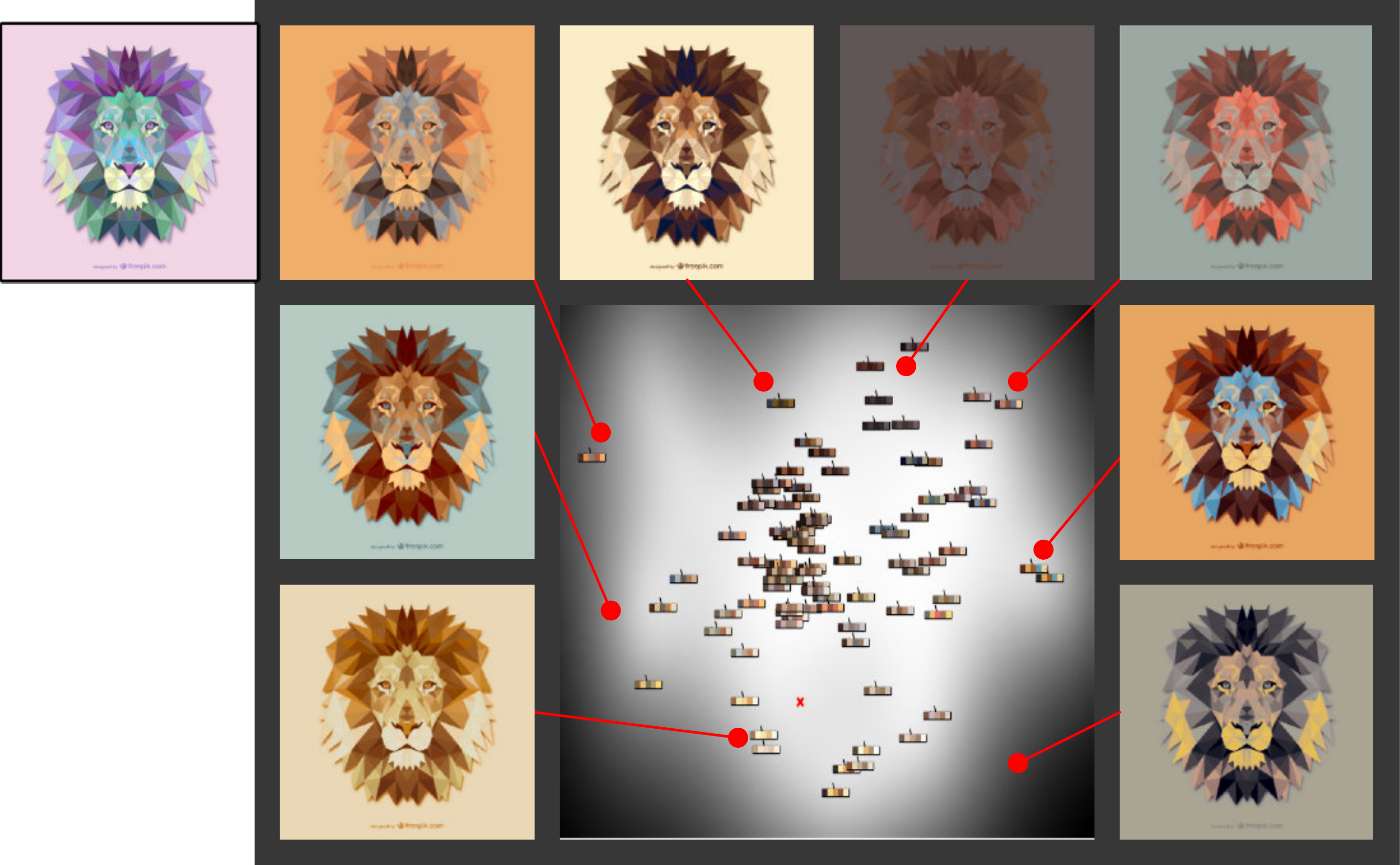} & \includegraphics[width=0.41\linewidth]{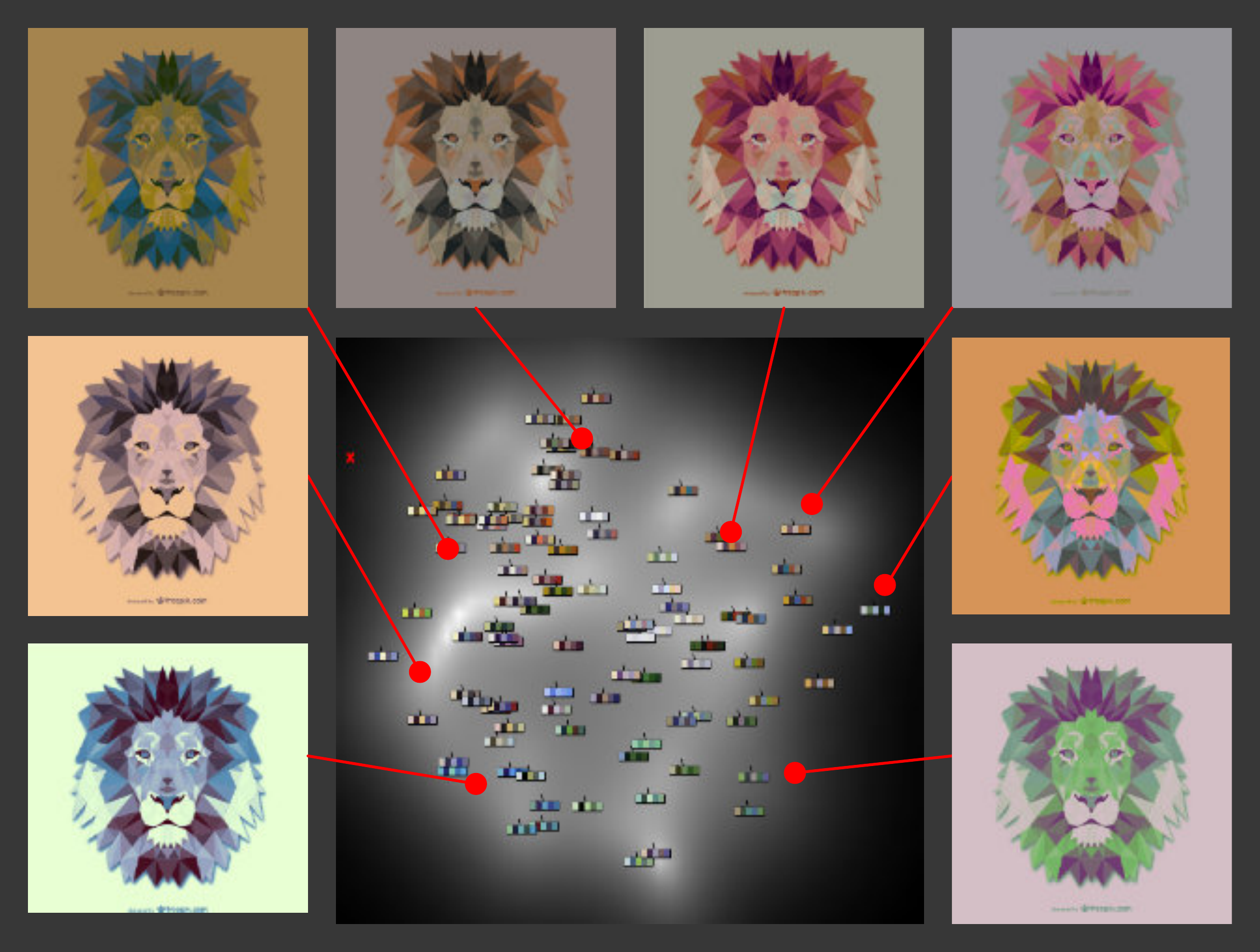} \\ 
\end{tabular} 
\caption[Exploration of the palette manifold.]{Exploration of the palette manifold learned from the ``Raphael'' set (Left) and the ``Luce'' set (Right). Red dots indicate the locations where the color palettes were selected for recolorization.  The original image (top-left corner) was designed by Freepik.com. }\label{fig:explorer_results}
\vspace{-10pt}
\end{figure*}

Photo stylization applications  like Lightroom offer users with a set of pre-defined filters and let them combine these filters to achieve a certain effect. It is, however, required that users know exactly what they want (a blurry photo in black and white, for instance). 
Inexperienced users often do not have such a clear picture of the final results but rather want to see different results and pick one that they prefer. We present an application called Photo-style Explorer, in which the user is allowed to freely surf through an infinite space of color styles and intuitively refine the result just by dragging the mouse. Figure \ref{fig:explorer_interface} shows the interface of the explorer and Figure \ref{fig:explorer_results} displays the representative results for two palette manifolds ``Luce'' and ``Raphael''. The colorization technique involved here was identical to the one described in Section~\ref{sec:photo_recolorization} except that we cached intermediate values to make the application works in real-time. 

\vspace{-5pt}
\subsection{Adaptive Color Palette}
\begin{figure}[ht]
\centering
\includegraphics[width=0.7\linewidth]{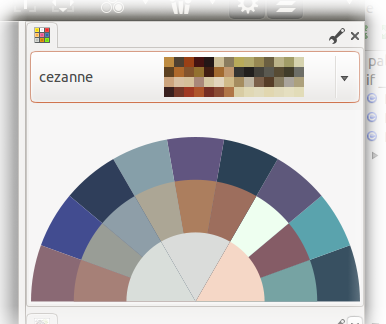}
\caption{SmartPalette interface. Please refer to the accompanying video for painting with real-time suggested palettes. The user may choose different artistic styles via the combo box at the top.   
}\label{fig:smartpalette}
\vspace{-10pt}
\end{figure}

To present the color combinations suggested by our method to the user, we designed a novel color picking interface, called ``SmartPalette'' interface. SmartPalette has been integrated into MyPaint~\cite{mypaint2015}, an open-source painting software written in Python. As shown in Figure~\ref{fig:smartpalette},
we chose the half-disk design that allows the colors to be shown in the order of their density in a patch.  Each 60-degree slice (1/3 of the half-disk) presents a set of colors (7 colors in our implementation). We sorted the colors by their brightness so that they will not ``jump'' around every time a new prediction is made. The colors were arranged into the slice in left-to-right and outer-inner orders. Different color suggestions were given by changing the order of the given colors (see the accompanying video). Different from previous methods~\cite{o2011color,lin2013probabilistic}, which considered only suggestion over a whole painting, our tool adaptively checked the regions being painted and performed palette prediction. We sampled the pixels from a 200x200 pixels patch using the same method described in Section \ref{sec:dataset}. The result was instantly presented to the user. The whole prediction process took less than half a second on an ordinary PC (Core 2 Duo 2.4 ghz, 8GB memory). 

\begin{figure}[ht]
\centering
\includegraphics[width=0.9\linewidth]{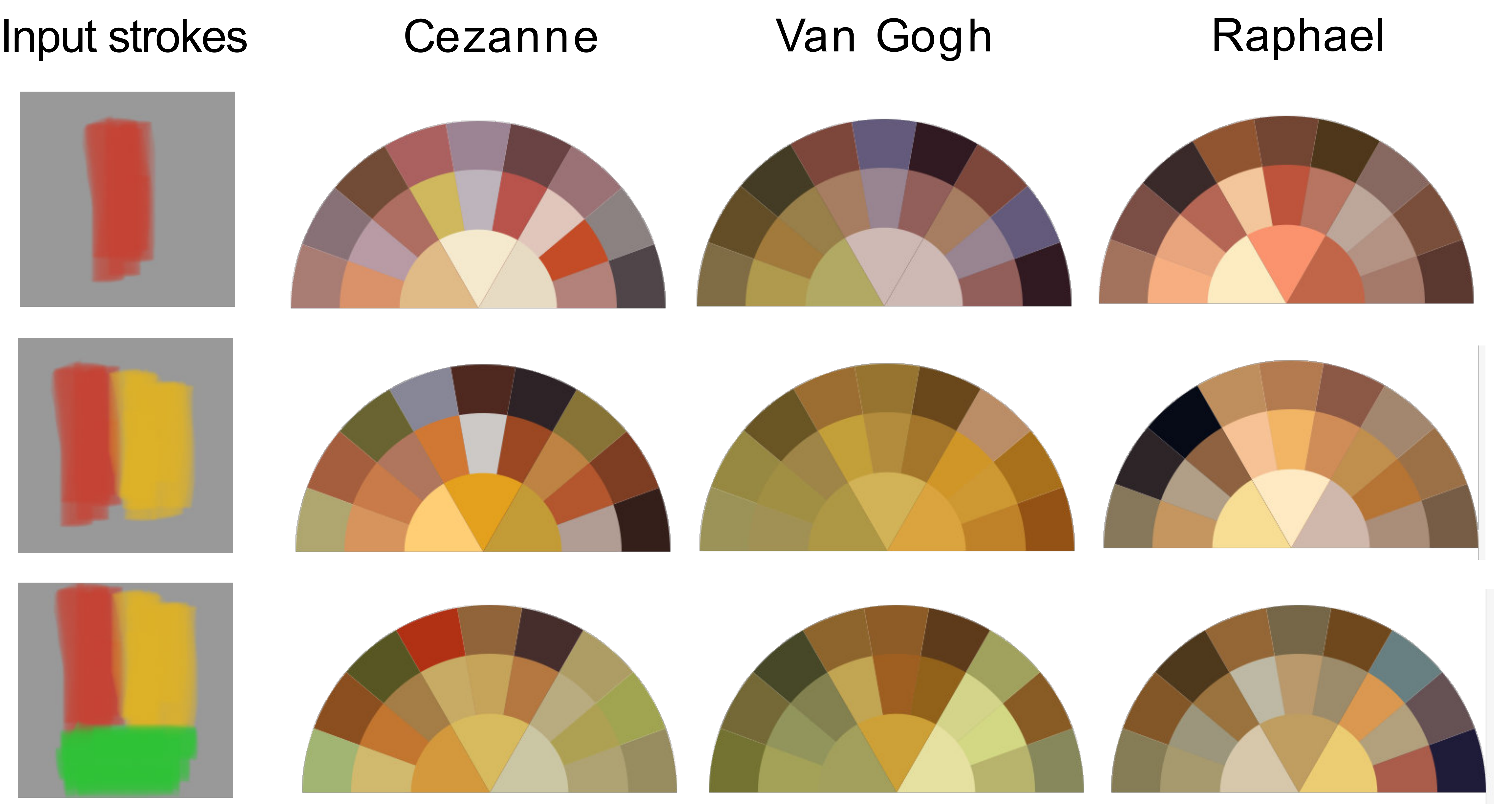}
\caption{Sample suggestions from SmartPalette. The left-most column shows the query strokes. The remaining columns are the suggestions from different artist sets.}\label{fig:suggestions}
\vspace{-15pt}
\end{figure}

Since our learning method is unsupervised, it is easy for professional painters/designers to train their own SmartPalette and use it to improve productivity. With GPLVM, the system does not require a lot of examples. Tens of example paintings are sufficient. Similarly, a design team can also make use of SmartPalette to constrain the use of colors so that their products will have consistent color themes. Figure \ref{fig:suggestions} shows different suggestions provided by SmartPalette in an incremental manner.  Here we used models trained on 3 different sets: Cezanne, Gogh and Raphael sets.
First, the user commits an initial brush stroke (red), which results in suggestions with reds and other predictably compatible colors. By changing the artist, the user can obtain different suggestions that bear the style of that artist. Second, the user commits a second brush stroke (yellow), the palettes now contain both reds and yellows and other colors. The interaction is repeated throughout the design process. One might notice that the variability of colors in Raphael set is not very large. This is because Raphael was a painter in the Classical period and he mainly used black, red, dirt yellow in his paintings.
\paragraph*{\textbf{Multi-style painting and User Study}}
To see how SmartPalette works in real-life scenarios, we asked two users with art backgrounds to paint using our color suggestion tool. The traditional HSV wheel was also shown for selecting initial colors. The users were provided a line sketch along with a grey-scale photo of the same scene for reference. Color suggestions were based on the Cezanne, Luce, or Raphael sets. Figure \ref{fig:colorization} (bottom) shows the paintings produced with the palettes from the three artists. Note that, the paintings bear resemblance in color styles with the painting sets. For instance, the painting of User 1 that used suggestions from Luce set has ocean blue shadows like in original paintings. The paintings made with suggestions from Raphael set are less vibrant and are more brownish.

To quantitatively assess our Adaptive Color Palette, we replicated the above experiment with 4 users with experiences in digital painting, and recorded relevant information. There was no time limit for the test but it normally took about 8 minutes for each user to complete the test. We recorded the number of times the user chose a color from either our palette or the HSV wheel, the number of times these colors were actually painted and the number of times the users manually requested for color suggestions from our method. \huy{We did not run the test with just the HSV wheel as it might be subject to unaccountable factors such as painting skill and test timing. For instance, professional designers may perform just as good with or without our tool.  However, the tool will still be useful if it learns from their own painting collection}.

The results show that, the users preferred the colors from our method 81\% of the times on average. 84\% of the total painting time was done with colors from our suggestions. Each user manually requested for suggestions for 1.4 times per minute during the test. We also asked the users to rate different aspects of our method (from 1 to 5, the higher the better). The users agreed that they will use our tool frequently (4.3/5), the tool was easy to use (4.0/5) and most people can learn to use it quickly (4.3/5). Some users suggested that SmartPalette should display similar and novel colors in separate sections and the software should keep a history of suggested palettes.

 \vspace{-10pt}
\section{Evaluation and Results}\label{sec:evals}
We evaluated our sorting algorithm by comparing it to a baseline method and then quantitatively assessed the predictive performance of the learning models, which operate on our feature vector.

\subsection{Palette Sort Comparisons}\label{smartpalette:sec:evalsort}
\begin{figure*}[ht]
\centering
\includegraphics[width=0.9\linewidth]{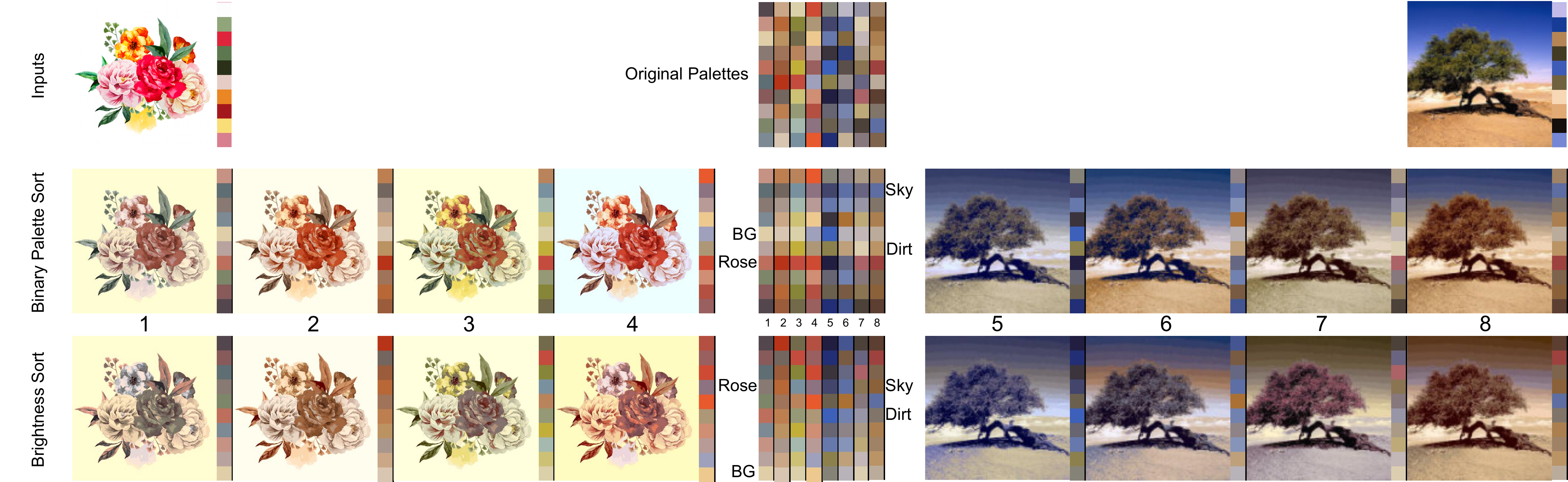}
\caption[SmartPalette: Ordering performance for multiple palettes.]{Ordering performance is demonstrated with the recolorization results. We show the corresponding palette datasets at the middle. The palettes sorted with our method are annotated with semantic labels (Rose, Sky, Dirt and Background \---\ BG), indicating the objects that get the colors from the rows.}\label{fig:alignmentcomparison}
\vspace{-15pt}
\end{figure*}
We chose simple palette ordering methods, which are based on brightness and hue, as the baselines to evaluate our method. In the experiments, we conveniently call these methods \emph{Brightness Sort} and \emph{Hue Sort}. As colors in paintings often follow certain shading rules, brightness and hue can be basic cues for aligning them~\cite{mollica2013color}. 
These simple methods, however, failed when the palettes to be ordered do not have strong contrast or hue, which is the case of our datasets. 
\begin{figure}[ht]
\centering
\includegraphics[width=0.8\linewidth]{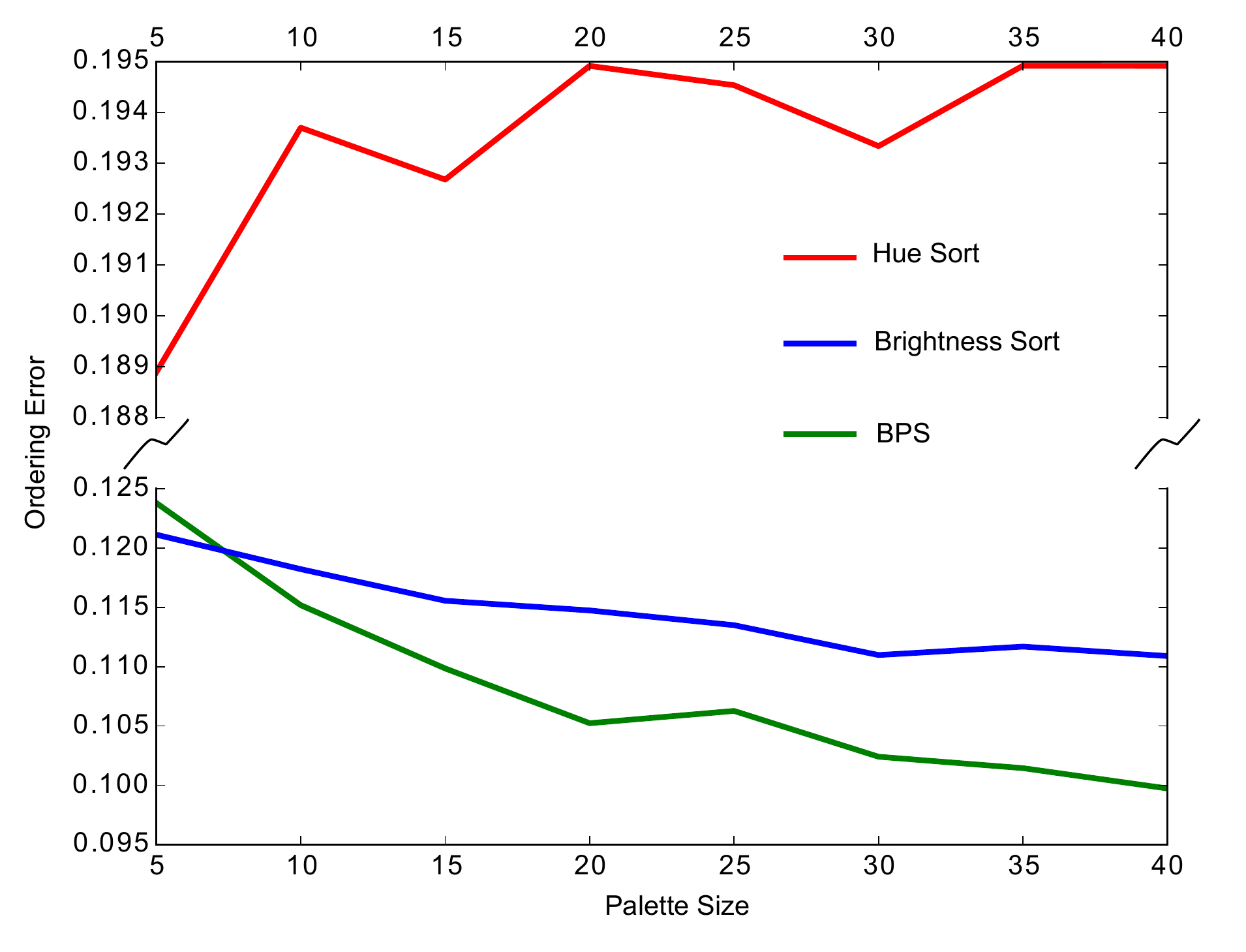}
\caption[Palette ordering with different methods.]{Palette ordering with different methods: Hue Sort, Brightness Sort and BPS.}\label{fig:vary_palette_size}
\vspace{10pt}
\end{figure}
\vspace{-10pt}
\subsubsection{Ordering Performance}\label{smartpalette:sec:eval_ordering}
To measure the ordering performance, we computed the mean distance between the colors in two consecutive palettes, averaged over all palettes in a dataset. Since we are interested in ordering colors in the palettes so that similar palettes have better correspondences, we pre-ordered the palettes with KPCA before the experiment (see Section~\ref{smartpalette:sec:binarysortalgo}). In  Figure~\ref{fig:vary_palette_size}, we plot the distance against the palette size. For each palette size $K$, we randomly selected a dataset of size $20 \times K$ (20 palettes) out of a dataset of size $20 \times 40$ and then ordered it with 3 methods: BPS, Brightness Sort and Hue Sort. Each experiment was repeated 5 times and the results were averaged over all 17 sets. As shown in the figure, Hue Sort has the worst error rates, comparing to BPS and Brightness Sort. Thus, we will not further consider it in the later sections. On the other hand, Brightness Sort only produces comparable performance to BPS when $K \leq 8$. As $K$ increases, BPS increasingly outperforms Brightness Sort. Note that, the color distance appears to inversely proportional with the size of palette. This is because there are more correspondences in a more complete dataset. To visually assess the ordering performance, we carried out a toy experiment as follows.

First, we chose two source images and extracted one 10-color palette per image. Next, we selected a set of 8 palettes from the Prendergast and the Luce set and then used them for recolorization. To make the correspondences between the extracted palettes and the selected palettes more visible to the reader, we used MHD to choose palettes that are similar to the extracted palettes.
Next, we sorted the palette set with BPS and Brightness Sort.  The motivation for having two input images is to demonstrate the localization property of our method. 
As shown in Figure \ref{fig:alignmentcomparison}, there are two distinctive groups of palettes in the palette set ([1-4] and [5-8]), each group resembles an input palette. As Binary Palette Sort aligns the palettes from local to global, we can observe clear correspondences for both groups of palettes. For example, the blues, representing the sky, and the reds, representing the rose, have consistent indices across the palettes. This demonstrates a case of \emph{context localization}, an natural behavior of our method. As discussed in Section~\ref{sec:reordering}, our sorting algorithm is able to localize the painting contexts embedded in a palette dataset. By assuming that similar palettes should describe similar objects or regions, our algorithm partitions and sorts these palettes together. The algorithm creates correspondences in color between nearby palettes, and thus helps preserving the context of each color locally.

To see how different color palettes may affect recolorization, we applied the sorted palettes to recolor the same source image (the same as in Figure~\ref{fig:alignmentcomparison}). Colors with the same indices were applied to the same regions in the source image (the first color went to the background, the second color went to the small flowers, etc.). The sorted palettes were matched with the palette extracted from the source image by using the same technique discussed in \ref{smartpalette:sec:binarysortalgo}. We used the recolorization technique introduced in ~\cite{levin2004colorization}. Note that, the recolorization method used in this section is different from the one in Section~\ref{sec:photo_recolorization}. As the purpose was only to visualize the differences between color palettes and we had only one palette to recolorize the image, the technique in ~\cite{levin2004colorization} was more appropriate. As a result of correct alignment, our recolorization results remain consistent across different palettes. For example, the rose stays red and the sky stays blue in all the images. In contrast, the results produced with Brightness Sort appear to be quite inconsistent as the blues and reds keep switching regions.

 To further confirm the superiority of our method, we conducted a user study to 
compare the consistency among the results produced by our method and Brightness Sort. We followed the same procedure discussed above but with 10 source images. We chose the source images in a way that ensures distinctiveness between colors in each source image. The full set of test images can be found in the supplementary documents for this paper. We generated a set of 6 images for each pair of sorting method and source image. The resulting sets for the two methods and the source image were displayed side-by-side for comparison and the order of appearance is counter-balanced. 15 users were asked to pick a set that has better consistency. \huy{Concretely, we instructed the user to look at each image set as a whole and to examine the color consistency within each set. We provided examples of consistent and inconsistent sets if the user is still unclear about the task. We also informed the user that they should not care about the aesthetic quality of the images as it is irrelevant to the test. The results show that the users preferred our method 127/150 of the times $(\chi^2=74.9, p<0.0001)$}.

\vspace{-5pt}
\subsubsection{Interpolation Performance}\label{smartpalette:sec:eval_interpolation}
\begin{figure*}[ht]
\centering
\includegraphics[width=0.95\linewidth]{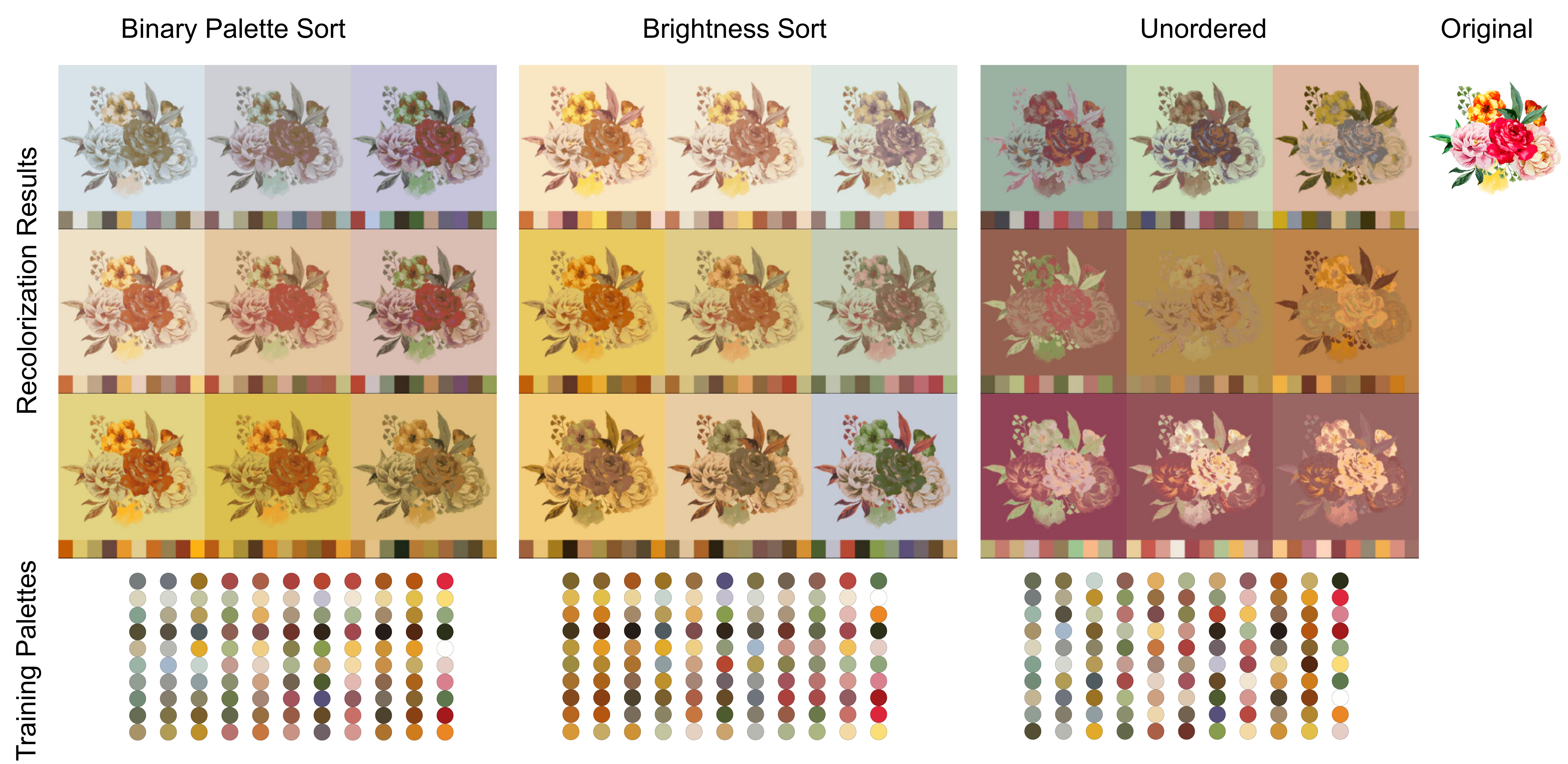}
\caption{Palettes sampled from GPLVM's latent space were applied to the same input image in Figure~\ref{fig:alignmentcomparison}. The training palettes are at the bottom, each column is a 10-color palette.}\label{fig:gplvm_sampling}
\end{figure*}

We demonstrate the interpolation performance of BPS by comparing it to the baseline method.
We performed interpolation on the low dimensional space induced by PCA or GPLVM and then projected the interpolated points back to the palette space. We also visualized the color palettes by applying the interpolated palettes to the same source image, as in Section~\ref{smartpalette:sec:eval_ordering}. Figure \ref{fig:gplvm_sampling} shows recolorization results with the palettes sampled along an evenly spaced grid from the latent space induced by GPLVM. The sampled palettes are placed under each recolorized image. Under each set of recolorization results, we show the corresponding palettes used for training the model.  When the palettes are correctly ordered with Binary Palette Sort, we obtain better palette interpolations as the colors stay ``clean'' and consistent across the interpolated palettes \---\ e.g., the rose stays red in all cases. 
With brightness-based ordering, the interpolated colors turn brown as they are just the averages of the misaligned palettes (top row, Brightness Sort). In the last row of Brightness Sort, the roses are colored green, as a result of interpolating between misaligned colors (reds and greens). The results in the Unordered case appear to be unusable as the colors  jump arbitrarily from one place to another. The results produced with PCA are shown in the supplementary.



\subsection{Quantitative Comparison}
%
%
\begin{figure}[H]
\centering
\includegraphics[width=0.9\linewidth]{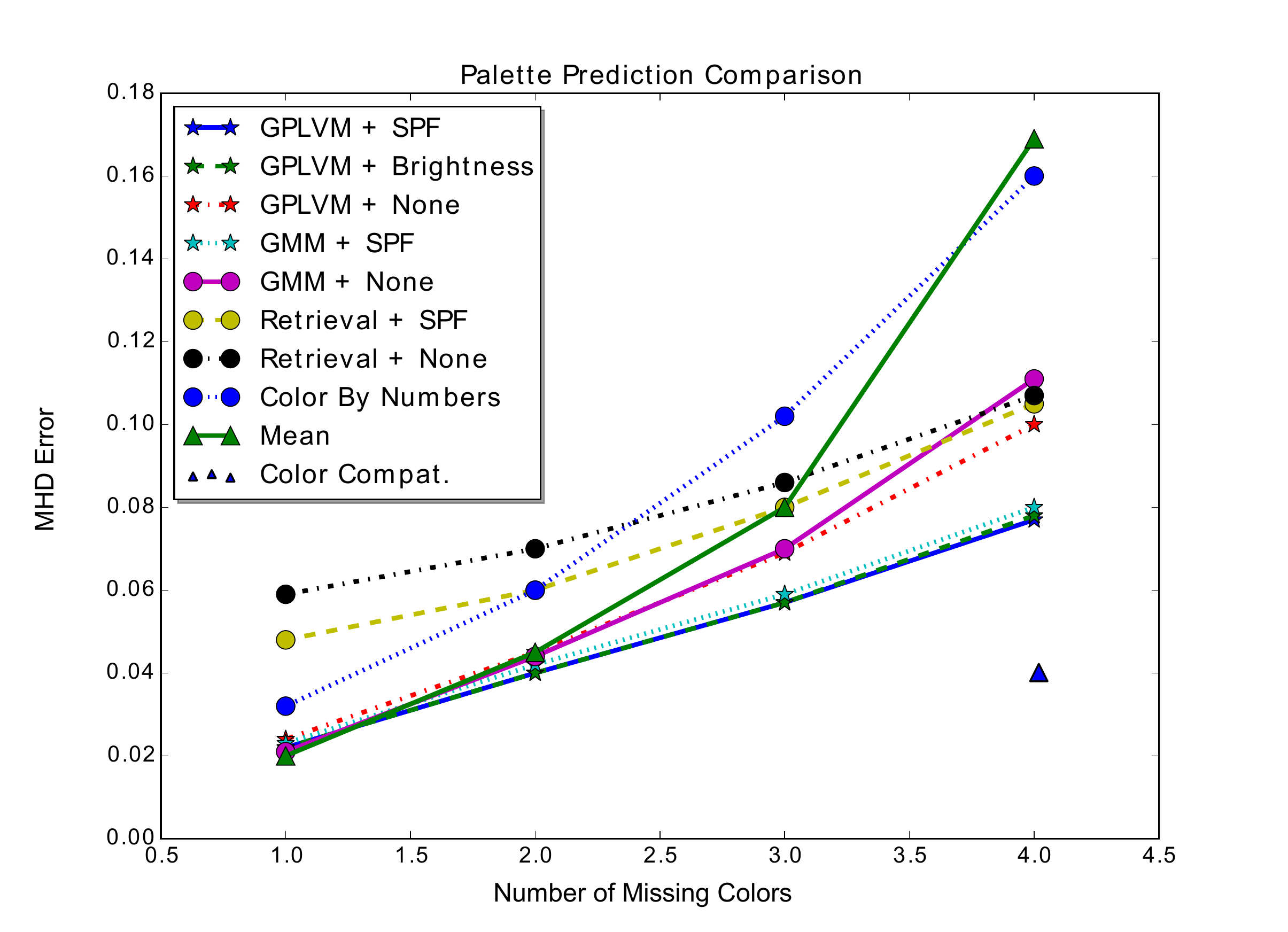}
\caption{\huy{Comparing theme completion performance. Prediction error are measured by the average MHD distance between predicted themes and ground-truth themes. The smaller, the better}}\label{smartpalette:fig:palette_pred_comparison}
\vspace{-10pt}
\end{figure}
We assessed the quality of the proposed model quantitatively by applying it to the task of color prediction. 
In this experiment, datasets were divided into training and testing sets with the training ratio of 0.6 (60\% of the whole set used for training). We conducted the experiment on 3 sets: Renoir, De Gas and Chase and reported the averaged results. Each set contained roughly from 500 to 600 5-color palettes.  Given 4, 3, 2 and 1 colors from an \emph{unordered} palette, the task was to predict/reconstruct the original palette (5 colors). Each experiment was repeated 5 times on random splits. We experimented with different methods for palette prediction. From now on, we refer to the vectorized ordered palettes produced by our Binary Palette Sort as Sorted Palette Features (SPF). 
\begin{enumerate}
\item GPLVM + SPF: GPLVM on SPF (introduced in Section \ref{sec:density}).
\item GPLVM + Brightness: GPLVM with Brightness Sort.
\item GPLVM + None: GPLVM on the original palette data.
\item GMM + SPF:  Gaussian Mixture Regression with SPF.
\item GMM + None: GMM on the original palette data.
\item Retrieval + SPF: We retrieved the most similar palette to the given colors. Since the palettes were already sorted, we used Euclidean distance to measure the similarity  between the given colors and colors with the same indices in the dataset.
\item Retrieval + None: This method is similar to the above method but the original (unordered) palette data was used.
\item Color Compatibility \cite{o2011color}: One of the first data-driven methods for color harmony discovery and palette rating prediction. Due to the slow implementation, we could only try it on the single color prediction case. The implementation was taken from the authors.
\item Color by Numbers \cite{lin2013probabilistic}: A method that relies on unitary, pairwise and global color functions connected by a factor graph to predict harmonious color combinations. The global function was taken from \cite{o2011color}. To apply the method to our data, we used raw patches extracted from the images as input to the system. We used the implementation provided by the authors of the paper. 
\item Mean: A simple method that takes the average of all provided colors as prediction. We used it as a baseline for other methods.

\end{enumerate}

\begin{figure}
\centering
\includegraphics[width=0.7\linewidth]{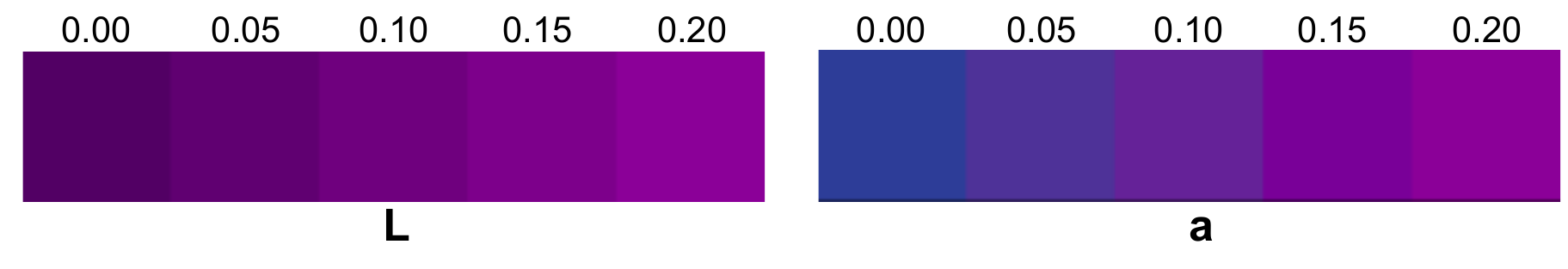}
\caption{Color scales for ``L'' and ``a'' channels in CIE*Lab space. The numbers are the relative distances to the first color.}\label{smartpalette:fig:color_scale}
\vspace{-10pt}
\end{figure}

For the case of GMM, GPLVM and Retrieval we first ran the Binary Palette Sort algorithm on the training set to obtain the SPF vectors. Then, for each given set of observed colors, we found a few similar palettes with MHD (3 palettes in our experiment) in the training set and then aligned the input set to these examples using the same technique used  to sort a pair of palette sets in Binary Palette Sort (see Section \ref{sec:reordering}). Even though the observed set might have missing colors, the matching algorithm still works as it only needs the pairwise distances between the colors.
The missing colors in the palettes were then predicted using the technique described in Section \ref{sec:palette_completion}. Since the predicted results were still unordered (except for the 1-color case), we again used MHD to measure the distance between the predicted theme and the ground-truth. Since MHD is simply an average distance between pairs of colors, we visualize the relative differences between colors in Figure~\ref{smartpalette:fig:color_scale}. A change of 0.05 in either ``L'' channel or ``a'' channel can produce visible differences. The reported numbers are the average distance between all predictions and ground-truths.

Fig. \ref{smartpalette:fig:palette_pred_comparison} shows the average error of the methods with different numbers of missing colors. One might notice that methods Color by Numbers (9) and Color Compatibility (8) are not suitable for the task of palette completion, since their error rates appear to be higher than other methods.
We run an independent t-test to test the difference between our method (1) and (9) and get $\text{t-statistic}=-2.47, \text{p-value}=0.02$, which means that our method is significantly better in predicting colors (smaller error rates). We also found via t-test that the learning-based methods with SPF (1) and (4) are significantly better than the retrieval-based methods (6) and (7). This result justifies our choice of using learning approaches for predicting color palettes. In contrast, the learning-based methods without SPF (3) and (5) are not significantly better than any of the retrieval-based methods. The Brightness Sort method, which was studied in Section \ref{smartpalette:sec:evalsort}, yields very similar results to our method in this test. This can be explained by the fact that brightness is the most important factor in visual data. In fact, many computer vision algorithms only need gray-scale data as the cue for predictions.
 
 GPLVM-based methods are generally better than GMM-based ones for both types of data (sorted and unordered). For this reason, we chose GPLVM for the applications introduced in Section \ref{sec:apps}. For both prediction methods, we observe a boost in performance when the SPF vectors are used. 
 It is worth noting that the mean predictor (Mean) works best for the case of 1 missing color. This can be explained by looking at our dataset. Since we gathered color themes from local patches, there were many patches that contain different shades of the same color. Thus, the missing colors in those cases are fairly similar to the average of the other colors in the same theme.  The t-test results, however, did not confirm the significance of the difference in error rate between (1), (3), (4), (5) and (10). In Section \ref{smartpalette:sec:evalsort}, we visually evaluated these methods.
 \vspace{-10pt}
\section{Limitations and Future Work}
\begin{figure}[ht]
\centering
\includegraphics[width=0.6\linewidth]{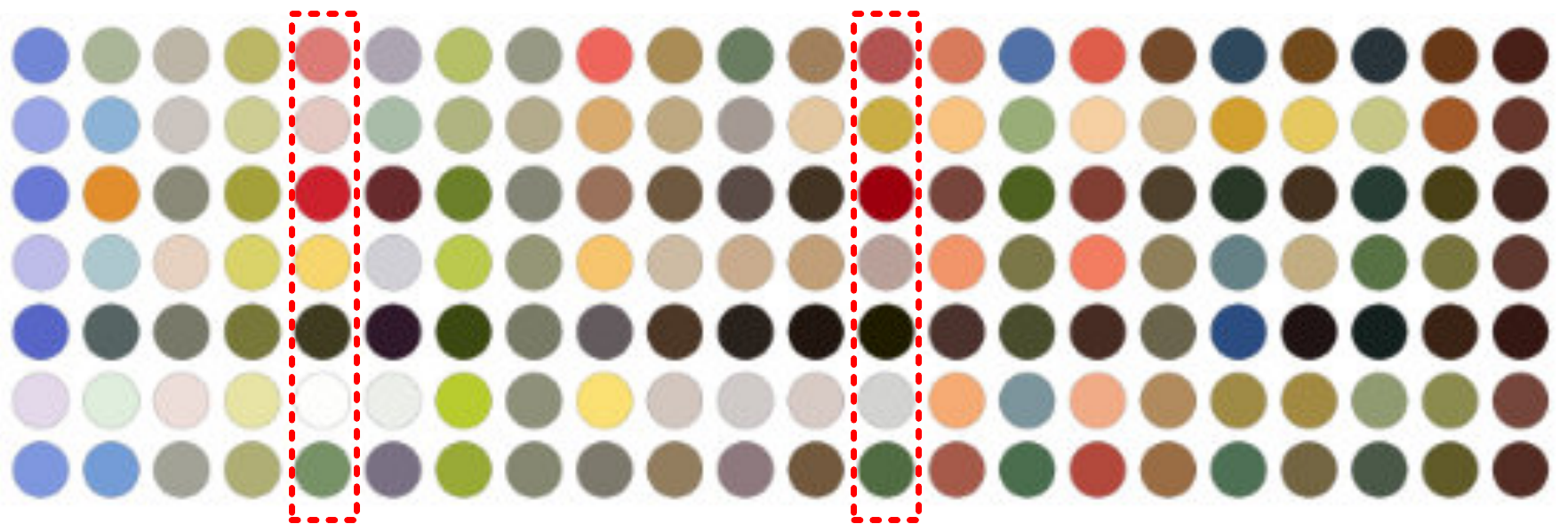}
\caption{A case when distant palettes are not perfectly aligned.}\label{fig:failed_alignment}
\vspace{-10pt}
\end{figure}

\begin{figure}[ht]
\centering
\includegraphics[width=0.9\linewidth]{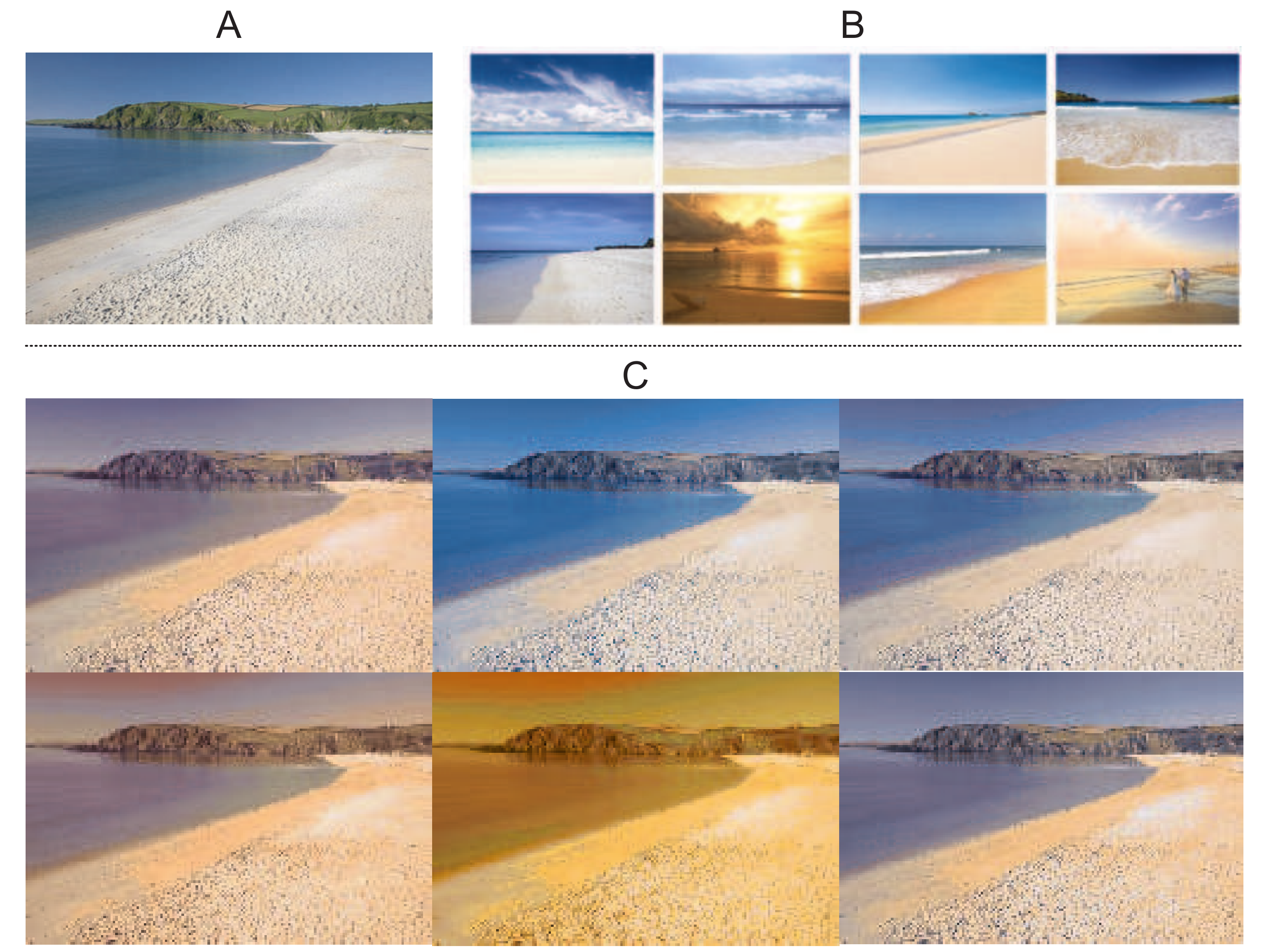}
\caption{\huy{Context-specific recolorization. (A) Original image. (B) Training set. (C) Various results.}}\label{fig:context_specific}
\vspace{-5pt}
\end{figure}

Our approach still suffers a number of limitations. First of all, our algorithm does not guarantee perfect alignment between distant palettes. This is due to the fact that we aligned nearby palettes first while distant palettes were implicitly aligned via set alignment (see Section~ \ref{sec:reordering}). In Figure \ref{fig:failed_alignment}, we intentionally include a palette and its brightness-shifted version (palettes in red boxes) into the set. Because the latter is shifted, the two palettes are separated from each other, which causes a slight misalignment (the yellow and the light pink swapped positions). A possible way to improve this is to replace the MHD kernel in the KPCA sorting step (Section \ref{sec:reordering}) with a more advanced kernel, which also considers the structure of the palettes instead of raw distances between colors. Another limitation of our work is, although the photo-style exploration can be potentially useful for real-world use cases, we have not yet evaluated it with user studies. Due to the subjectivity of aesthetic evaluation, it is not straightforward to judge the quality of the stylized results. 
In the future, we would like to study how our model would work on context-specific photo recolorization problems. That is, given a training set containing images of similar scenes such as beach, mountain, summer, etc. would it be possible to make use of our model to recolorize novel images in a consistent way. In Figure~\ref{fig:context_specific}, we show some preliminary results for a particular case of beach scene. We used the same technique discussed in Sec \ref{smartpalette:sec:photoexplorer}. The results look quite promising because areas in the source image have been recolorized with meaningful colors (the sand turns yellow and sky turns indigo).

\huy{Another possible future work is to study how different segmentation affect the recolorization results. Since our method suggests color palettes based on input colors extracted from segments, a bad segmentation can result in less distinctive results. For example, if the input palette is just shades of the same color (due to a color-based segmentation, for example), the output is likely to be the same because there are indeed many similar palettes in the training set. Thus, in our recolorization pipeline, a good segmentation should partition image into objects and meaningful regions. A similar effect can be seen when comparing the results in Fig. \ref{fig:colorization} and Fig. \ref{fig:explorer_results}. Due to the real-time constraint, the photo-explorer use input palettes extracted globally, which are more distinctive, thus produces more varying results at the cost of local shades. In contrast, the results in Fig.  \ref{fig:colorization} is produced with local input palettes, resulting in richer shades. Depending on the application, one might want to choose one approach or combine them to achieve desirable results.}

\vspace{-5pt}
\section{Conclusion}
We have introduced a novel method for interpolating and summarizing palette data. Palette datasets are often available as unordered sets of colors, making it difficult to directly apply traditional methods to analyze the data. We designed an effective palette ordering method (Binary Palette Sort) that makes use of kernel-based dimensionality reduction to reorder colors in palettes in a meaningful way, allowing us to apply state-of-the-art interpolation techniques on palette data. The palette density provided a mean to develop numerous interesting applications such as real-time adaptive palette, photo-style exploration, enriched photo recolorization. We have conducted both quantitative and qualitative experiments to assess the performance of our method, and favorable results were obtained. In the future, we would like to study in depth how the user interacts with SmartPalette and Photo-style Explorer, possibly through crowd-sourcing and online survey.

\ifCLASSOPTIONcaptionsoff
  \newpage
\fi



\bibliographystyle{IEEEtran}
\bibliography{main}
%



%
\vspace{-20pt}
\begin{IEEEbiography}[{\includegraphics[width=1in,height=1.25in,clip,keepaspectratio]{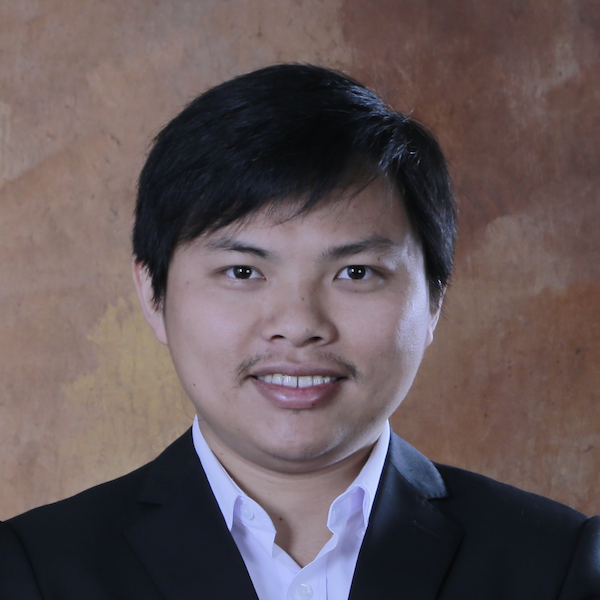}}]{Huy Q. Phan}
is an postdoctoral researcher at the University of Bath. He received his B.Sc in Computer Science from the Vietnam National University in 2009 and attained a PhD from the City University of Hong Kong in 2016, under supervisions of Prof. Hongbo Fu and Prof. Antoni B. Chan. In 2015, he was a research intern at Adobe Systems in San Jose city.
\end{IEEEbiography}
\vspace{-20pt}
\begin{IEEEbiography}[{\includegraphics[width=1in,height=1.25in,clip,keepaspectratio]{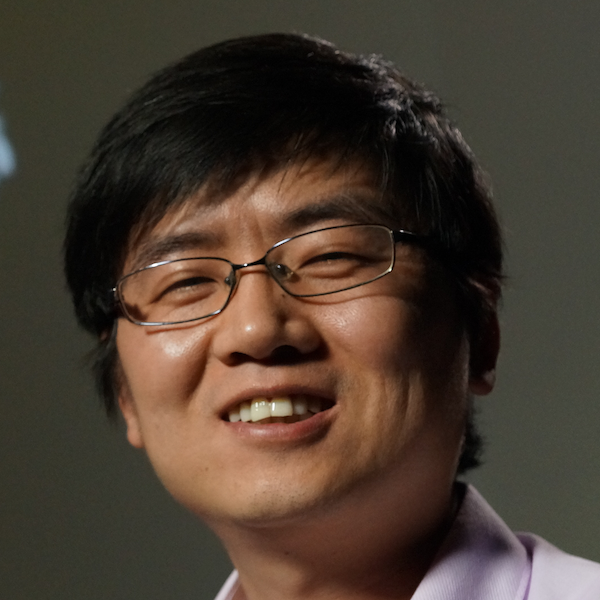}}]{Hongbo Fu}
is an Associate Professor in the School of Creative Media, City University of Hong Kong. He received the PhD degree in computer science from the Hong Kong University of Science and Technology in 2007 and the BS degree in information sciences from Peking University, China, in 2002. His primary research interests fall in the fields of computer graphics and human computer interaction. He was the Program Chair or Co-chair of CAD/Graphics 2013, SIGGRAPH Asia 2013 (Emerging Technologies), SIGGRAPH Asia 2014 (Workshops) and CAD/Graphics 2015. He is the Conference Chair for SIGGRAPH Asia 2016. He also serves as an Associate Editor of The Visual Computer (TVC), Computers \& Graphics (C\&G), and Computer Graphics Forum (CGF).
\end{IEEEbiography}
\vspace{-20pt}
\begin{IEEEbiography}[{\includegraphics[width=1in,height=1.25in,clip,keepaspectratio]{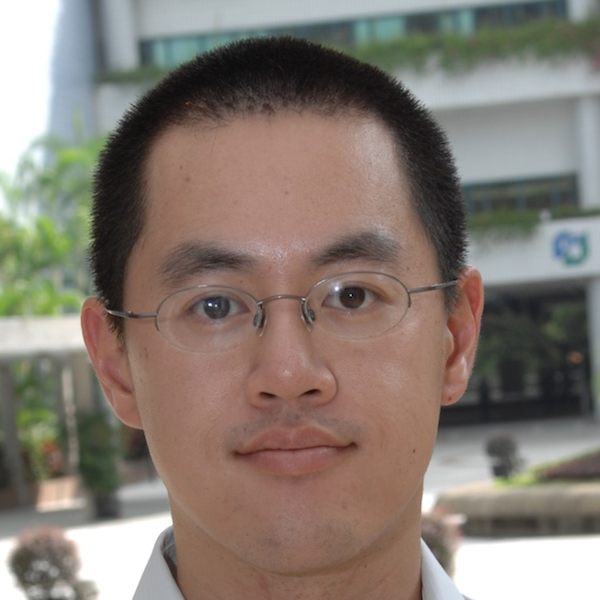}}]{Antoni B. Chan}
received the B.S. and M.Eng. degrees in electrical engineering from Cornell University, Ithaca, NY, USA, in 2000 and 2001, respectively, and the Ph.D. degree in electrical and computer engineering from University of California at San Diego (UCSD), La Jolla, CA, USA, in 2008. He was a Visiting Scientist with the Vision and Image Analysis Laboratory, Cornell University, from 2001 to 2003, and a Post-Doctoral Researcher with the Statistical Visual Computing Laboratory, UCSD, in 2009. In 2009 he joined the Department of Computer Science, City University of Hong Kong, Hong Kong, and is currently an Associate Professor. His research interests include computer vision, machine learning, pattern recognition, eye-gaze analysis, and music analysis. Dr. Chan received the National Science Foundation Integrative Graduate Education and Research Training Fellowship from 2006 to 2008, and an Early Career Award from the Research Grants Council of the Hong Kong Special Administrative Region, China, in 2012.
\end{IEEEbiography}









\end{document}